\documentclass[10pt,twocolumn,letterpaper]{article}

 \usepackage[pagenumbers]{iccv} % To force page numbers, e.g. for an arXiv version

\usepackage{bm}
\usepackage{booktabs}
\usepackage{multirow}
\usepackage{times}
\usepackage{microtype}
\usepackage{epsfig}
\usepackage{caption}
\usepackage{float}
\usepackage{placeins}
\usepackage{color, colortbl}
\usepackage{stfloats}
\usepackage{enumitem}
\usepackage{tabularx}
\usepackage{xstring}
\usepackage{xspace}
\usepackage{url}
\usepackage{subcaption}       
\usepackage[dvipsnames]{xcolor}
\usepackage{ulem}

\newcommand\blfootnote[1]{%
  \begingroup
  \renewcommand\thefootnote{}\footnote{#1}%
  \addtocounter{footnote}{-1}%
  \endgroup
}

\definecolor{gtable}{rgb}{0.0, 0.5, 0.0}

\newcommand{\ra}[1]{\renewcommand{\arraystretch}{#1}}

\makeatletter

\newcommand{\Rmnum}[1]{\expandafter\@slowromancap\romannumeral #1@}
\makeatother

\usepackage[utf8]{inputenc}
\usepackage{amssymb}
\usepackage{bbding}
\usepackage{pifont}
\usepackage{wasysym}
\usepackage{utfsym}
\usepackage{fontawesome}
\definecolor{gray}{gray}{0.8} % 可以调整灰度值来改变颜色深浅

\newcommand{\green}[1]{{\color[HTML]{5F8D4E}#1}}

\newcommand{\gray}[1]{{\color[HTML]{B7B7B7}#1}}

\definecolor{gain}{HTML}{34a853}  %
\newcommand{\gain}[1]{\textcolor{gain}{#1}}
\definecolor{lost}{HTML}{ea4335}  %

\newcommand{\good}[2]{{#1} {{\gain{#2}}}}

\definecolor{iccvblue}{rgb}{0.21,0.49,0.74}
\usepackage[pagebackref,breaklinks,colorlinks,allcolors=iccvblue]{hyperref}

\def\confName{ICCV}
\def\confYear{2025}

\title{CreatiLayout: Siamese Multimodal Diffusion Transformer for \underline{Creati}ve \underline{Layout}-to-Image Generation}

\author{
    Hui Zhang$^{1,2,3}$\qquad
    Dexiang Hong$^{3}$\qquad
    Yitong Wang$^{3}$\qquad
    Jie Shao$^{3}$\qquad 
    Xinglong Wu$^{3}$\qquad \\
    Zuxuan Wu$^{1,2,\dagger}$\qquad
    Yu-Gang Jiang$^{1,2}$
    \\
    $^{1}$Institute of Trustworthy Embodied AI, Fudan University \\
    $^{2}$Shanghai Collaborative Innovation Center of Intelligent Visual Computing  
    \\
    $^{3}$Bytedance Intelligent Creation
    \vspace{-8pt}
}

\begin{document}

\twocolumn[{
\maketitle
\begin{center}
    \vspace{-18pt}
    \includegraphics[width=0.95\linewidth]{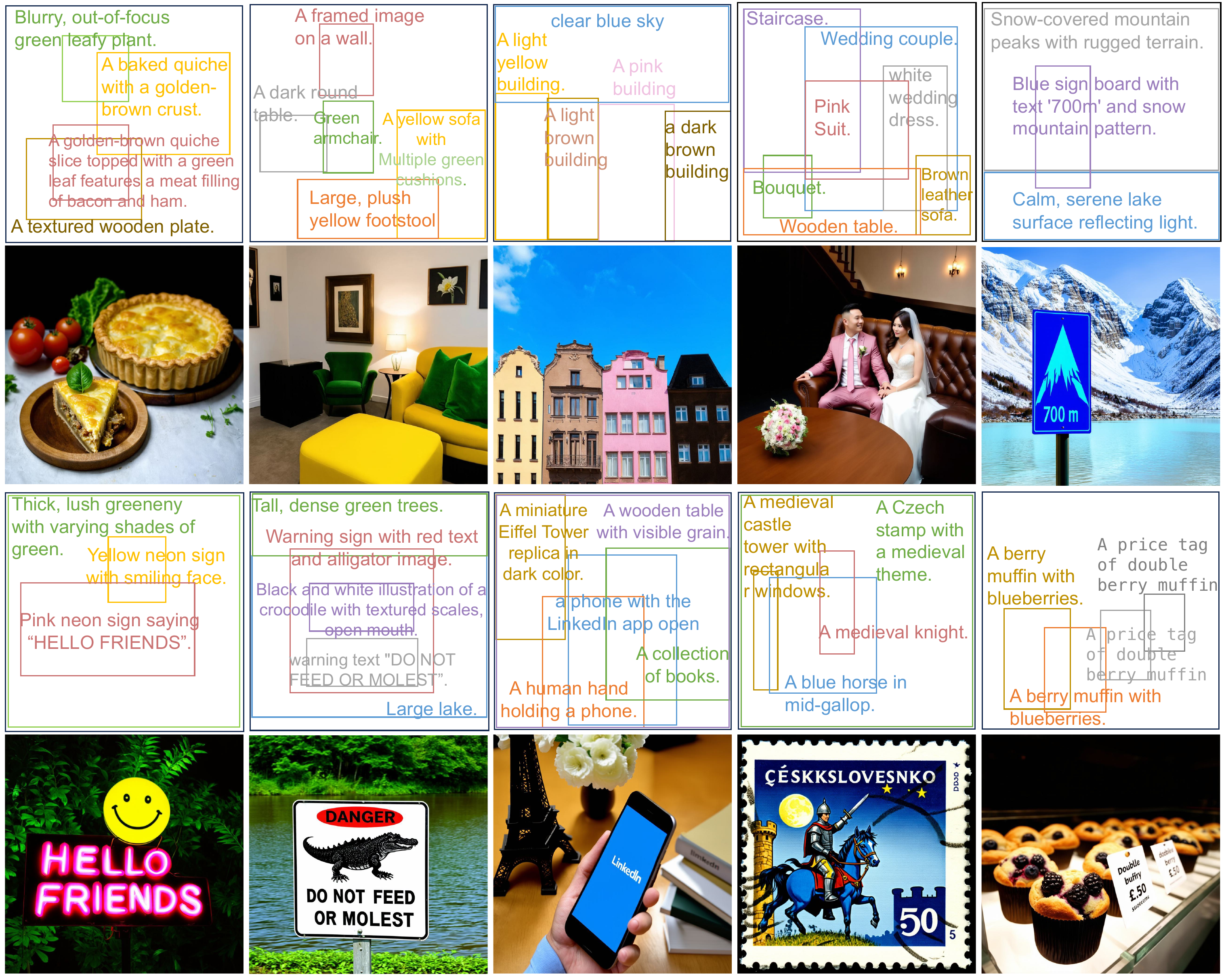}
    \vspace{-5pt}
    \captionof{figure}{We present a novel approach to empower MM-DiTs for layout-to-image generation, achieving high-quality and fine-grained controllable generation, as evidenced by the precise rendering of complex attributes (\eg color, texture, shape, and text).}
    \label{fig:teaser}
    \vspace{-0.5em}
\end{center}
}]

\maketitle
\vspace{-1.0em}
\begin{abstract}
\blfootnote{$^{{\dagger}}$ Corresponding author.}
Diffusion models have been recognized for their ability to generate images that are not only visually appealing but also of high artistic quality.
As a result, Layout-to-Image (L2I) generation has been proposed to leverage region-specific positions and descriptions to enable more precise and controllable generation.
However, previous methods primarily focus on UNet-based models (\eg SD1.5 and SDXL), and limited effort has explored  Multimodal Diffusion Transformers (MM-DiTs), which have demonstrated powerful image generation capabilities.
Enabling MM-DiT for layout-to-image generation seems straightforward but is challenging due to the complexity of how layout is introduced, integrated, and balanced among multiple modalities.
To this end, we explore various network variants to efficiently incorporate layout guidance into MM-DiT, and ultimately present SiamLayout.
To inherit the advantages of MM-DiT, we use a separate set of network weights to process the layout, treating it as equally important as the image and text modalities.
Meanwhile, to alleviate the competition among modalities, we decouple the image-layout interaction into a siamese branch alongside the image-text one and fuse them in the later stage.
Moreover, we contribute a large-scale layout dataset, named LayoutSAM, which includes 2.7 million image-text pairs and 10.7 million entities. Each entity is annotated with a bounding box and a detailed description.  
We further construct the LayoutSAM-Eval benchmark as a comprehensive tool for evaluating the L2I generation quality.
Finally, we introduce the Layout Designer, which taps into the potential of large language models in layout planning, transforming them into experts in layout generation and optimization.
These components form \textbf{CreatiLayout}---a systematic solution that integrates the layout model, dataset, and planner for \underline{creati}ve \underline{layout}-to-image generation.
Code, model, and dataset:\url{https://creatilayout.github.io/}.

\end{abstract}    
\vspace{-1.5em}
\section{Introduction}
\label{sec:intro}
Text-to-image (T2I) generation~\cite{rombach2022stablediffusion,podell2023sdxl,chen2023pixartalpha,li2024hunyuandit} has been widely applied and deeply ingrained in various fields thanks to the rapid advancement of diffusion models~\cite{ho2020ddpm,song2021ddim}.
To achieve more controllable generation, Layout-to-image (L2I) has been proposed to generate images based on layout conditions consisting of spatial location and description of entities.

Recently, multimodal diffusion transformers (MM-DiTs)~\cite{esser2024sd3,sd3.5,flux,li2024playground} have taken text-to-image generation to the next level. These models treat text as an independent modality equally important as the image and utilize MM-Attention~\cite{esser2024sd3} instead of cross-attention for interaction between modalities, thus enhancing prompt following.
However, previous layout-to-image methods~\cite{li2023gligen,wang2024instancediffusion,zhou2024migc} mainly fall into UNet-based architectures~\cite{rombach2022stablediffusion,podell2023sdxl} and achieve layout control by introducing extra image-layout fusion modules between the image's self-attention and the image-text cross-attention.
While enabling MM-DiT for layout-to-image generation seems straightforward, it is challenging due to the complexity of how layout is introduced, integrated, and balanced among multiple modalities. 
To this end, \textbf{\textit{there is a pressing need to tailor a layout integration network for MM-DiTs}}, fully unleashing their capabilities for high-quality and precisely controllable generation.

To address this issue, we explore various network variants and ultimately propose SiamLayout.
Firstly, \textbf{\textit{we treat layout as an independent modality}}, equally important as the image and text modalities. More specifically, we employ a separate set of transformer parameters to process the layout modality. During the forward process, the layout modality interacts with other modalities via MM-Attention and maintains self-updates.
Secondly, \textbf{\textit{we decouple the interactions among the three modalities into two siamese branches: image-layout and image-text MM-Attentions}}. By independently guiding the image with text and layout and then fusing them at a later stage, we alleviate competition among modalities and strengthen the guidance from the layout.

\begin{table}[t]
\centering
\tabcolsep=0.08cm
\ra{1.1}
\scalebox{0.7}{
\begin{tabular}{@{}l|ccc|c@{}}
\toprule
Layout Dataset    & COCO~\cite{lin2014coco} & Instance~\cite{wang2024instancediffusion} & Ranni~\cite{feng2024ranni}& \textbf{LayoutSAM} \\ \midrule
Spatial Location   & \Checkmark & \Checkmark & \Checkmark & \Checkmark \\
Detailed Global Description & \textcolor{gray}{\XSolidBrush} & \textcolor{gray}{\XSolidBrush} & \Checkmark & \Checkmark \\
Detailed Region Description & \textcolor{gray}{\XSolidBrush} & \Checkmark & \textcolor{gray}{\XSolidBrush} & \Checkmark \\
Open-Set Entity         & \textcolor{gray}{\XSolidBrush} & \textcolor{gray}{\XSolidBrush} & \Checkmark & \Checkmark \\ \bottomrule
\end{tabular}}
\caption{Compared to previous layout datasets, LayoutSAM consists of open-set entities with fine-grained annotations.}
\label{tab:dataset comparison}
\vspace{-2.0em}
\end{table}

To this end, a high-quality layout dataset composed of image-text pairs and entity annotations is crucial for training the layout-to-image model.
As shown in ~\cref{tab:dataset comparison}, the closed-set and coarse-grained nature of existing layout datasets may limit the model's ability to generate complex attributes (\eg color, shape, texture).
Thus, we construct an automated annotation pipeline and contribute a large-scale layout dataset derived from the SAM dataset~\cite{kirillov2023sam}, named LayoutSAM. It includes 2.7M image-text pairs and 10.7M entities. Each entity includes a spatial position (\ie bounding box) and a region description.
The descriptions of images and entities are fine-grained, with an average length of 95.41 tokens and 15.07 tokens, respectively.
We further introduce the LayoutSAM-Eval benchmark to provide a comprehensive tool for evaluating layout-to-image generation quality.

To support diverse user inputs rather than just bounding boxes of entities, we turn a large language model into a layout planner named LayoutDesigner.
This model can convert and optimize various user inputs such as center points, masks, scribbles, or even a rough idea, into a harmonious and aesthetically pleasing layout.

These three components---\textbf{SiamLayout}, \textbf{LayoutSAM}, and \textbf{LayoutDesigner}---jointly form the systematic layout-to-image solution \textbf{CreatiLayout}, which integrates the layout model, dataset, and planner. Through comprehensive evaluations on LayoutSAM-Eval and COCO benchmarks, SiamLayout outperforms other variants and previous SOTA models by a clear margin, especially in generating entities with complex attributes, as illustrated in \cref{fig:teaser}. For layout planning, LayoutDesigner shows more comprehensive and specialized capabilities compared to baseline LLMs.

\vspace{-0.5em}
\section{Related Work}
\label{sec:related work}
\vspace{-0.5em}
\paragraph{Text-to-Image Generation.}
Text-to-image generation~\cite{rombach2022stablediffusion,podell2023sdxl,betker2023dalle3,saharia2022imagen,chen2023pixartalpha,li2024hunyuandit,peebles2023dit,bao2023u-vit,gao2024lumina,xing2024survey,tian2025unigen} has emerged as a promising application due to its impressive capabilities.
Recently, studies such as SD3~\cite{esser2024sd3}, SD3.5~\cite{sd3.5}, FLUX.1~\cite{flux}, and Playground-v3~\cite{liu2024playgroundv3} have advanced the multimodal Diffusion Transformer architecture (MM-DiT), elevating text-to-image generation to the next level. MM-DiT significantly enhances text understanding by treating text as an independent modality, equally important as the image, and replacing traditional cross-attention with MM-Attention for modal interaction.

\vspace{-1.2em}
\paragraph{Layout-to-Image Generation.}
To achieve more precise and controllable generation, layout-to-image generation~\cite{li2023gligen,wang2024instancediffusion,zhou2024migc,feng2024ranni,dahary2024beyourself,gong2024check,shirakawa2024noisecollage,phung2024attentionrefocusing,lee2024groundit,wu2024ifadapter,jia2024ssmg,yang2023law-diffusion,zheng2023layoutdiffusion,xing2023vidiff,xing2024aid,xing2024simda,xue2023freestyle,hoe2024interactdiffusion,ma2025hico,chen2023reasonyourlayout,nie2024blobgen,li2025magicmotion} has been proposed to generate images based on layout guidance, which includes several entities. Each entity comprises a spatial location and a region description.
However, previous methods primarily focused on UNet-based architectures, and enabling MM-DiT for L2I generation is challenging due to the complexity of introducing, integrating, and balancing the layout among multiple modalities.
In this paper, we focus on exploring winner solutions for incorporating layout into MM-DiT to unlock its power and enable precise and controllable generation.

\vspace{-1.2em}
\paragraph{Layout Datasets.}
Layout datasets typically consist of image-text pairs with entity annotations.
A common type~\cite{li2023gligen,zheng2023layoutdiffusion,yang2023law-diffusion} originates from COCO~\cite{lin2014coco}, featuring images with global descriptions and entities marked by bounding boxes and brief descriptions.
Although some effort~\cite{wang2024instancediffusion,shirakawa2024noisecollage,jia2024ssmg} expands descriptions using Large Language Models (LLMs) or Vision-Language Models (VLMs), they are still limited due to the close-set nature. Ranni~\cite{feng2024ranni} collects large-scale text-image pairs from LAION~\cite{schuhmann2022laion5b} and WebVision~\cite{li2017webvision}, moving towards an open-set layout dataset.
However, entity descriptions remain coarse-grained and lack complex attributes.
In this paper, we present an annotation pipeline and introduce a large-scale layout dataset containing 2.7M image-text pairs and 10.7M detailed entity annotations.

\vspace{-1.2em}
\paragraph{Large Language Model for Layout Generation.}
Layout generation~\cite{guerreiro2025layoutflow,horita2024retrieval-augmented,weng2024desigen} refers to the multimodal task of creating layouts for flyers, magazines, UI interfaces, or natural images.
Some studies~\cite{chen2023textdiffuser2,feng2024layoutgpt,qu2023layoutllm,cho2024visualprogramming,lianllm-grounded,wu2024self-correct,yang2024rpg-master,phung2024attentionrefocusing,feng2024ranni,omost} have explored using LLMs~\cite{achiam2023gpt4-turbo,team2024gemma-paper,dubey2024llama3.1-paper,hu2024minicpm-paper,hui2024qwen2.5-paper} to generate layouts based on textual descriptions, which then guide the generation of images. In this paper, we further enhance the capabilities of LLMs for generation and optimization as well as supporting user input of different granularities.
\section{Methodology}
\label{sec:method}

\subsection{Preliminaries}
\label{subsec:preliminaries}

\paragraph{Latent Diffusion Models~(LDMs).}
Latent diffusion models perform the diffusion process in the latent space, which consists of a VAE~\cite{kingma2013vae}, text encoders, and either a UNet-based or transformer-based noise prediction model $\epsilon_{\theta}$. The VAE encoder $\mathcal{E}$ encodes images $\mathbf{x}$ into the latent space $\mathbf{z}$, while the VAE decoder $\mathcal{D}$ reconstructs the latent back into images. The text encoders $\boldsymbol{\tau}$, such as CLIP~\cite{radford2021clip} and T5~\cite{raffel2020T5}, project tokenized text prompts into text embeddings $\mathbf{y}$. The training objective is to minimize the following LDM loss:
{
\setlength{\abovedisplayskip}{2pt}
\setlength{\belowdisplayskip}{2pt}
\begin{equation}
\begin{aligned}
\mathcal{L}_{LDM}=\mathbb{E}_{\mathbf{z}\sim\mathcal{E}(\mathbf{x}),\mathbf{y},\epsilon\sim\mathcal{N}(0,1),t}\left[\left\|\epsilon-\epsilon_\theta\left(\mathbf{z}_t,t,\mathbf{y}\right)\right\|_2^2\right],
\end{aligned}
\label{eq:ldm loss}
\end{equation}
}
where $t$ is time step uniformly sampled from $\{1,\ldots,T\}$. The latent $\mathbf{z}_t$ is obtained by adding noise to $\mathbf{z}_0$, with the noise $\epsilon$ sampled from the standard normal distribution $\mathbf{\mathcal{N}}(\mathbf{0},\mathbf{I})$.

\vspace{-1.0em}
\paragraph{Multimodal Diffusion Transformer.}
SD3/3.5~\cite{esser2024sd3,sd3.5}, FLUX.1~\cite{flux}, and Playground-v3~\cite{liu2024playgroundv3} instantiate noise prediction using MM-DiT, which uses two independent transformers to handle text and image embeddings separately. Unlike previous diffusion models that process different modalities through cross-attention, MM-DiT concatenates the embeddings of the image and text for the self-attention operation, referred to as MM-Attention:
{
\setlength{\abovedisplayskip}{2pt}
\setlength{\belowdisplayskip}{2pt}
\begin{equation}
\begin{aligned}
\mathbf{z}, \mathbf{y} = \text{Self-Attention}([\mathbf{z}, \mathbf{y}]). 
\end{aligned}
\label{eq:mm attention}
\end{equation}
}
MM-DiT treats image and text as equally important modalities to improve prompt following~\cite{esser2024sd3}.
In this paper, we explore incorporating layout into MM-DiTs, unleashing their potential for high-quality and precise L2I generation.

\begin{figure*}
  \centering
  \includegraphics[width=0.96\linewidth]{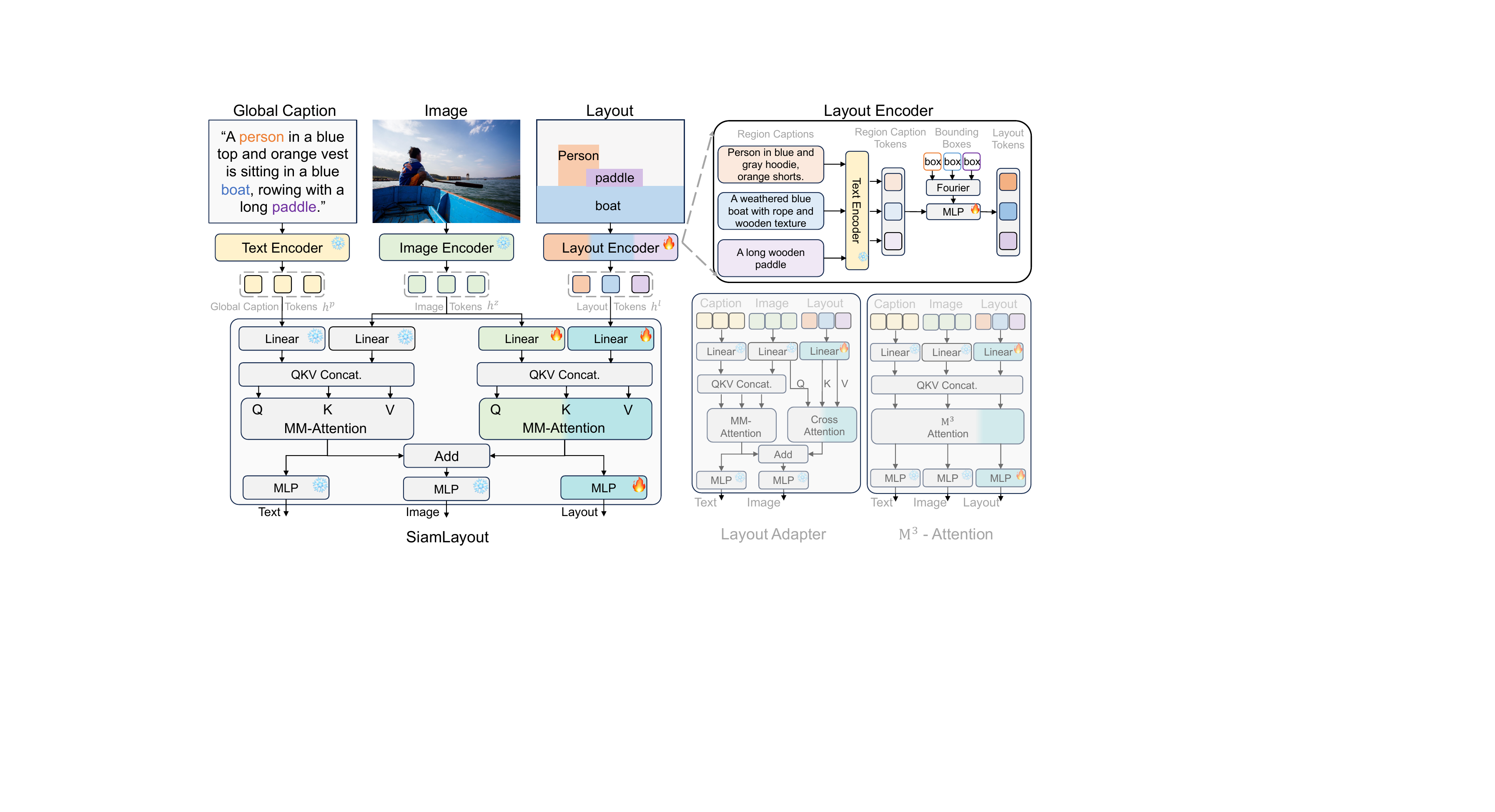}
  \vspace{-0.5em}
  \caption{\textbf{An overview of the proposed pipeline}. Layout tokens are derived from the layout encoder based on spatial locations and region descriptions. SiamLayout employs separate transformer parameters to process the layout, treating it as an equally important modality as the image and text.
  Layout and text guide the image independently through siamese branches, and are then fused in the later stage.
  We experiment with two additional network variants that incorporate layout via cross-attention and $\mathrm{M}^3$-Attention. SiamLayout works best.}
  \label{fig:architecture}
  \vspace{-1.5em}
\end{figure*}

\subsection{Layout-to-Image Generation}
\label{subsec:layout siamese}

\paragraph{Problem Definition.}
Layout-to-image generation aims at precise and controllable image generation based on the instruction $\boldsymbol{I}$, which consists of a global-wise prompt condition $\boldsymbol{p}$ and a region-wise layout condition $\boldsymbol{l}$, denoted as:
{
\setlength{\abovedisplayskip}{2pt}
\setlength{\belowdisplayskip}{2pt}
\begin{equation}
\begin{aligned}
\text{Instruction: }\boldsymbol{I}=(\boldsymbol{p},\boldsymbol{l}).
\end{aligned}
\label{eq:instruction}
\end{equation}
}
The layout condition includes information for $N$ entities $\boldsymbol{e}$, each consisting of two parts: region caption $\boldsymbol{c}$ and spatial location $\boldsymbol{b}$, denoted as:
{
\setlength{\abovedisplayskip}{2pt}
\setlength{\belowdisplayskip}{2pt}
\begin{equation}
\begin{aligned}
\text{Layout Condition: }\boldsymbol{l}=[\boldsymbol{e}_1,\cdots,\boldsymbol{e}_N], &\mathrm{with}\\
\end{aligned}
\label{eq:layout condition}
\end{equation}
\vspace{-1.0em}
\begin{equation}
\begin{aligned}
\text{Entity: }\boldsymbol{e}_n=(c_n,b_n).
\end{aligned}
\label{eq:entity}
\vspace{-0.2em}
\end{equation}
}
In this work, we use bounding boxes to represent spatial locations, consisting of the coordinates of the top-left and bottom-right corners.

\vspace{-1.0em}
\paragraph{Tokenize Different Modalities.}
Image tokens $\boldsymbol{h}^z$ are derived by patchifying the latent $\mathbf{z}$, and tokens $\boldsymbol{h}^p$ of global caption $\boldsymbol{p}$ are obtained from the text encoder $\boldsymbol{\tau}$, denoted as $\boldsymbol{h}^p=\boldsymbol{\tau}(\boldsymbol{p})$. 
We denote the layout tokens as \( \boldsymbol{h}^l = [h_1^l, \cdots, h_N^l] \). Inspired by GLIGEN~\cite{li2023gligen}, each \( h^l_i \) is obtained from the layout encoder in ~\cref{fig:architecture}:
{
\setlength{\abovedisplayskip}{2pt}
\setlength{\belowdisplayskip}{2pt}
\begin{equation}
\begin{aligned}
h^l_i=\mathrm{MLP}([\boldsymbol{\tau}(c_i),\mathrm{Fourier}(b_i)])
\end{aligned}
\label{eq:layout tokens}
\end{equation}
}
Fourier refers to the Fourier embedding~\cite{mildenhall2021fourier}, [·, ·] denotes concatenation across the feature dimension, and MLP is a multi-layer perception.

\begin{figure}[t]
  \centering
  \includegraphics[width=0.95\linewidth]{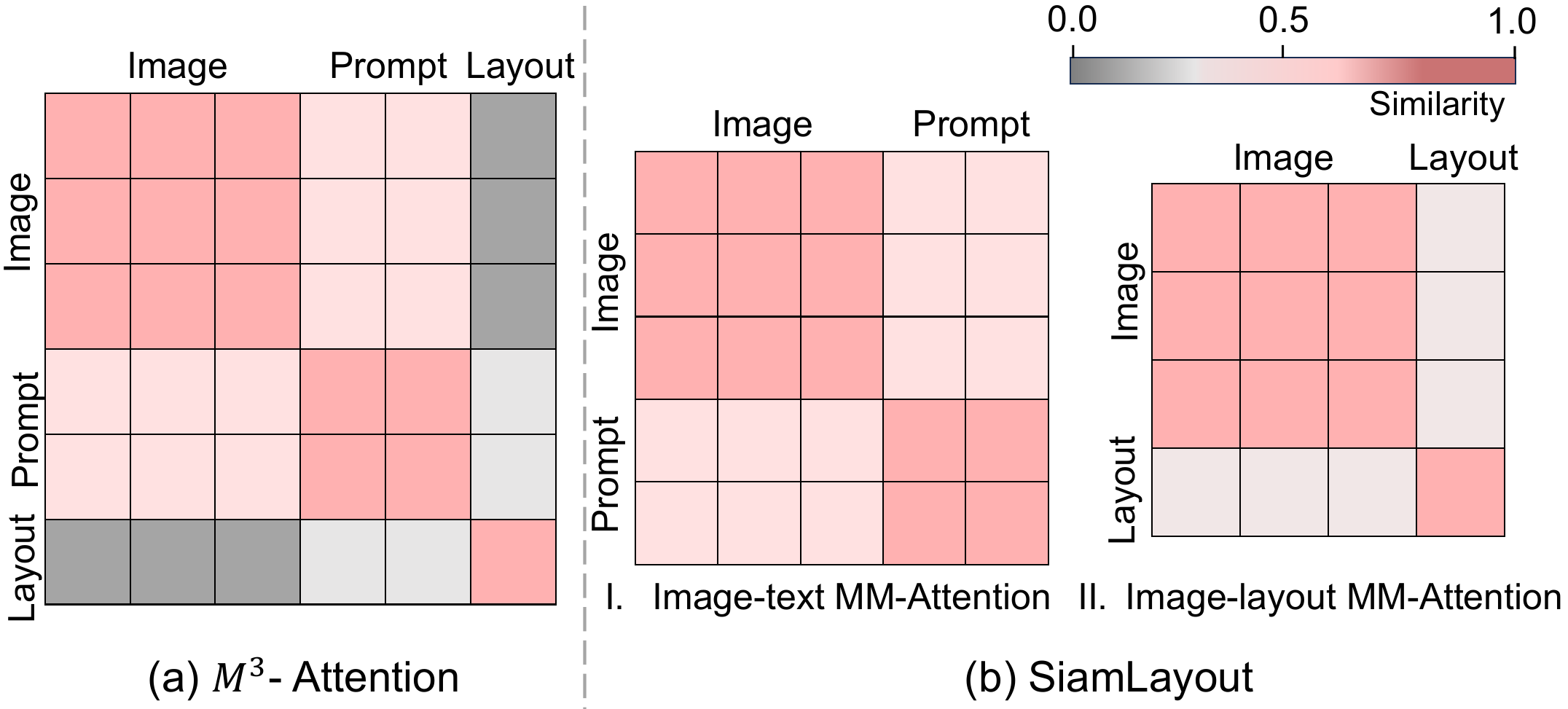}
  \caption{\textbf{Attention maps of $\mathbf{M}^3$-Attention and SiamLayout.} (a) The layout's influence on image generation is much lower compared to the text due to lower similarity. (b) SiamLayout decouples (a) into two independent MM-Attentions for image-text and image-layout, enabling equal guidance from both layout and text.}
  \label{fig:Modality_competition}
  \vspace{-1.5em}
\end{figure}

\vspace{-1.0em}
\paragraph{Layout Integration.}
MM-DiT~\cite{esser2024sd3} allows the two modalities to interact through the following MM-Attention:
\vspace{-0.2em}
\begin{small}
\begin{equation}
\begin{aligned}
{\boldsymbol{h}^{z}}^{\prime},{\boldsymbol{h}^{p}}^{\prime}=\mathrm{Attention}([\mathbf{Q}^{z},\mathbf{Q}^{p}],[\mathbf{K}^{z},\mathbf{K}^{p}],[\mathbf{V}^{z},\mathbf{V}^{p}]),
\end{aligned}
\label{eq:img-text mm attention}
\vspace{-0.2em}
\end{equation}
\end{small}
where [·, ·] denotes concatenation across the tokens dimension, $\mathbf{Q}^{z} = \boldsymbol{h}^{z}\mathbf{W}_q^z$, $\mathbf{K}^{p} = \boldsymbol{h}^{z}\mathbf{W}_k^z$, $\mathbf{V}^{z} = \boldsymbol{h}^{z}\mathbf{W}_v^z$; and $\mathbf{W}_q^z$, $\mathbf{W}_k^z$, $\mathbf{W}_v^z$ are the weight matrices for the query, key, and value linear projection layers for image tokens, respectively.
Tokens $\boldsymbol{h}^{p}$ of the global caption $\boldsymbol{p}$ are handled by the same paradigm as $\boldsymbol{h}^{z}$ but with their own weights.
${\boldsymbol{h}^{z}}^{\prime}$ and ${\boldsymbol{h}^{p}}^{\prime}$ are the image and caption tokens after interaction.
To this end, the next critical step is to incorporate the layout tokens $\boldsymbol{h}^l$.
We explore three variants of network designs that incorporate layout tokens, as shown in \cref{fig:architecture}. Here, we present an illustration of our method based on the SD3 model. For further details regarding FLUX's layout control mechanism, please refer to the supplementary material.
\begin{itemize}
\item[--] \textbf{\textit{Layout Adapter.}} 
Based on the fundamental idea of previous L2I methods~\cite{li2023gligen,ye2023ipadapter} introducing layout conditions in UNet-based architectures, we design extra image-layout cross-attention to incorporate layout into MM-DiT, defined as the Layout Adapter:
{
\setlength{\abovedisplayskip}{2pt}
\setlength{\belowdisplayskip}{3pt}
\begin{equation}
\begin{aligned}
\boldsymbol{h}_{\mathrm{adapter}}^z= {\boldsymbol{h}^{z}}^{\prime} + \mathrm{Attention}(\mathbf{Q}^{z},\mathbf{K}^{l},\mathbf{V}^{l}),
\end{aligned}
\label{eq:layout adapter}
\end{equation}
}
where $\mathbf{K}^{l} = \boldsymbol{h}^{l}\mathbf{W}_k^l$ and $\mathbf{V}^{l} = \boldsymbol{h}^{l}\mathbf{W}_v^l$ are the key and values matrices from the layout tokens. $\mathbf{W}_k^l$ and $\mathbf{W}_v^l$ are the corresponding weight matrices. $\mathbf{Q}^{z}$ is the identical query as in \cref{eq:img-text mm attention}. 
Guidance from the layout is introduced into $\boldsymbol{h}_{\mathrm{adapter}}^z$ with fewer parameters. However, due to layout conditions not playing an equal role to the caption condition, it may diminish the impact of layout guidance on the generation process and lead to suboptimal results.
\item[--] \textbf{\textit{${M}^3$-Attention.}} To emphasize the importance of layout conditions, we extend the core philosophy of MM-DiT, treating layout as an independent modality equally important as the image and global caption. We employ a separate set of transformer parameters to process layout tokens and design a novel $\mathrm{M}^3$-Attention for the interaction among these three modalities:
\vspace{-0.4em}
\begin{equation}
\begin{small}
\begin{aligned}
\boldsymbol{h}_{m^3}^z,\boldsymbol{h}_{m^3}^p,\boldsymbol{h}_{m^3}^l=\mathrm{Attention}([\mathbf{Q}^{z},\mathbf{Q}^{p},\mathbf{Q}^{l}], \\
[\mathbf{K}^{z},\mathbf{K}^{p},\mathbf{K}^{l}],[\mathbf{V}^{z},\mathbf{V}^{p},\mathbf{V}^{l}]).
\end{aligned}
\label{eq:m3 attention}
\end{small}
\vspace{-0.4em}
\end{equation}
where $\mathbf{Q}^{l} = \boldsymbol{h}^{l}\mathbf{W}_q^l$, $\mathbf{K}^{l} = \boldsymbol{h}^{l}\mathbf{W}_k^l$ and $\mathbf{V}^{l} = \boldsymbol{h}^{l}\mathbf{W}_v^l$ are the query, key and values matrices from the layout tokens. $\mathbf{W}_q^l$, $\mathbf{W}_k^l$ and $\mathbf{W}_v^l$ are the corresponding independent weight matrices.
We concatenate these three modalities in the token dimension and then facilitate interaction among them through self-attention. During the generation process, the layout condition acts as an independent modality, constantly interacting with other modalities and maintaining self-updates.
Although this design seems promising, we found that in the attention map of $\mathrm{M}^3$-Attention, the similarity between layout and image is much lower than that between the caption and image (as shown in \cref{fig:Modality_competition} (a)), resulting in the layout having much less influence on the image compared to the global caption. We refer to this phenomenon as ``modality competition'', which can be attributed to the fact that layout, as a new modality, has not been pre-trained and aligned on large-scale paired datasets like the global caption and image.

\item[--] \textbf{\textit{SiamLayout.}} 
Finally, in order to retain the advantages of MM-Attention in multimodal interaction while mitigating competition among them, we propose a new layout fusion network named SiamLayout. As shown in \cref{fig:architecture}, we decouple $\mathrm{M}^3$-Attention into two isomorphic MM-Attention branches, \ie siamese branches, to handle image-text and image-layout interactions independently and simultaneously.
The MM-Attention between image and layout can be formally denoted as:
{
\setlength{\abovedisplayskip}{2pt}
\setlength{\belowdisplayskip}{3pt}
\begin{equation}
\begin{footnotesize}
\begin{aligned}
{\boldsymbol{h}^{z}}^{\prime\prime},{\boldsymbol{h}^{l}}^{\prime}=\mathrm{Attention}([{\mathbf{Q}^{z}}^{\prime},\mathbf{Q}^{l}],[{\mathbf{K}^{z}}^{\prime},\mathbf{K}^{l}],[{\mathbf{V}^{z}}^{\prime},\mathbf{V}^{l}]),
\end{aligned}
\label{eq:img-layout mm attention}
\end{footnotesize}
\end{equation}
}
where ${\mathbf{Q}^{z}}^{\prime}$,${\mathbf{K}^{z}}^{\prime}$ and ${\mathbf{V}^{z}}^{\prime}$ are the new query, key, and value matrices obtained from the image token, with weight matrices different from those in \cref{eq:img-text mm attention}. 
The final image tokens are the fusion of text-guided and layout-guided image tokens:$\boldsymbol{h}_{\mathrm{siamese}}^z={\boldsymbol{h}^{z}}^{\prime} + {\boldsymbol{h}^{z}}^{\prime\prime}$.
Since the guidance from the layout and global caption has been decoupled---being independent in space and parallel in time---the issue of modality competition is significantly mitigated. As shown in \cref{fig:Modality_competition} (b), in the attention map of image-layout MM-Attention, the similarity between layout and image increases continually during training as layout takes a dominant role here.
\end{itemize}
In this paper, SiamLayout is chosen as the network that incorporates layout into MM-DiTs, and our primary experiments are conducted on it.
In addition, we develop a LoRA-based variant of SiamLayout, which achieves comparable layout control accuracy in a more lightweight manner.

\vspace{-1.0em}
\paragraph{Training and Inference.}
We freeze the pre-trained model and only train the newly introduced parameters ${\theta}^{\prime}$ using the following loss function:
{
\setlength{\abovedisplayskip}{2pt}
\setlength{\belowdisplayskip}{3pt}
\begin{equation}
\begin{small}
\begin{aligned}
\mathcal{L}_{layout}=\mathbb{E}_{\mathbf{z},\mathbf{p},\boldsymbol{l},t,\epsilon\sim\mathcal{N}(0,1)}\left[\left\|\epsilon-\epsilon_{\{\theta,{\theta}^{\prime}\}}\left(\mathbf{z}_t,t,\mathbf{p},\boldsymbol{l}\right)\right\|_2^2\right].
\end{aligned}
\label{eq:training loss}
\end{small}
\end{equation}
}
Here, we employ two strategies to accelerate the convergence of the model:
\Rmnum{1}) Biased sampling of time steps: Since layout pertains to the structural content of images, which is primarily generated during the larger time steps, we sample time steps with a 70\% probability from a normal distribution $\mathcal{N}(0.7*T,T)$ and with a 30\% probability from $\mathcal{N}(0,T)$.
\Rmnum{2}) Region-aware loss: We enhance the model's focus on areas specified by the layout by assigning greater weight to the region loss $\mathcal{L}_{region}$ associated with the regions localized by the bounding boxes in the latent space. The updated loss: 
{
\setlength{\abovedisplayskip}{2pt}
\setlength{\belowdisplayskip}{3pt}
\begin{equation}
\begin{aligned}
\mathcal{L}_{layout}^{\prime}=\mathcal{L}_{layout}+{\lambda}_{region} \times \mathcal{L}_{region}, 
\end{aligned}
\label{eq:final training loss}
\end{equation}
}
where ${\lambda}_{region}$ modulates the importance of $\mathcal{L}_{region}$.
During the inference phase, we perform layout-conditioned denoising only in the first 30\% of the steps~\cite{li2023gligen,zhou2024migc}.

\subsection{Layout Dataset and Benchmark}
\label{subsec:layout dataset}

\begin{figure}[h]
 \vspace{-1.0em}
  \centering
  \includegraphics[width=1.0\linewidth]{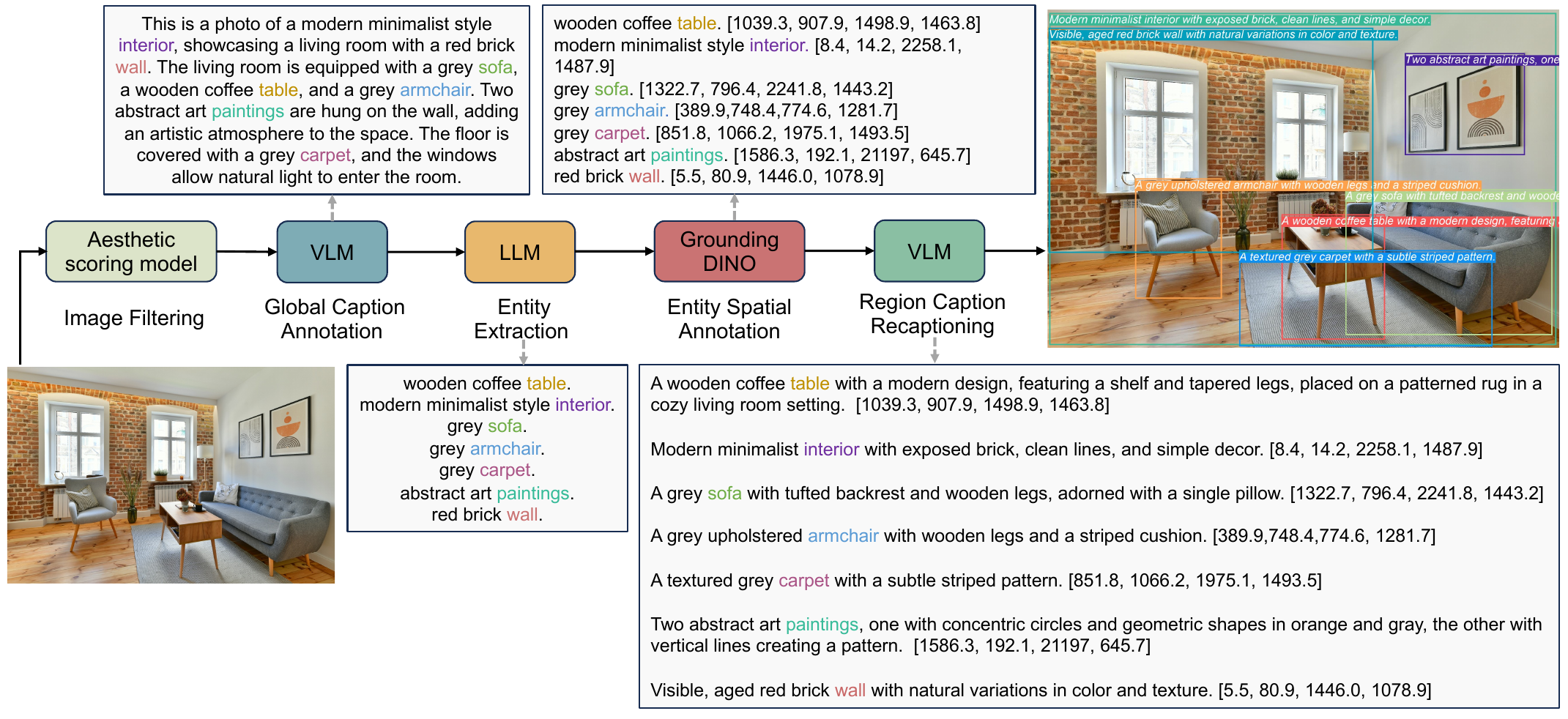}
  \vspace{-1.5em}
  \caption{An overview of the automatic annotation pipeline.}
  \label{fig:data_pipeline}
  \vspace{-2.0em}
\end{figure}

\paragraph{Layout Dataset.} As there is no large-scale and fine-grained layout dataset explicitly designed for layout-to-image generation, we collect 2.7 million image-text pairs with 10.7 million regional spatial-caption pairs derived from the SAM Dataset~\cite{kirillov2023sam}, named LayoutSAM.
We design automatic schemes and strict filtering rules to annotate layout and clean noisy data, with the following five parts:

\Rmnum{1}) Image Filtering: We employ the LAION-Aesthetics predictor~\cite{LAION-Aesthetics} to curate a high visual quality subset from SAM, selecting images in the top 50\% of aesthetic scores.

\Rmnum{2}) Global Caption Annotation: As the SAM dataset does not 
provide descriptions for each image, we generate detailed descriptions using a Vision-Language Model (VLM)~\cite{Qwen-VL-Captioner}. The average length of the captions is 95.41 tokens.

\Rmnum{3}) Entity Extraction: Existing SoTA open-set grounding models~\cite{liu2023grounding1,ren2024grounding1.5} prefer to detect entities through a list of short phrases rather than dense captions. Thus, we utilize a Large Language Model~\cite{llama3.1} to derive brief descriptions of main entities from dense captions via in-context learning. The average length of the brief descriptions is 2.08 tokens.

\Rmnum{4}) Entity Spatial Annotation: We use Grounding DINO~\cite{liu2023grounding1} to annotate bounding boxes of entities and design filtering rules to clean noisy data. Following previous work~\cite{zheng2023layoutdiffusion,yang2023law-diffusion}, we first filter out bounding boxes that occupy less than 2\% of the total image area, then only retain images with 3 to 10 bounding boxes.

\Rmnum{5}) Region Caption Recaptioning: At this point, we have the spatial locations and brief captions for each entity. We use a VLM~\cite{minicpm-v-2.6} to generate fine-grained descriptions with complex attributes for each entity based on its visual content and brief description. The average length of these detailed descriptions is 15.07 tokens.

\vspace{-1.0em}
\paragraph{Layout-to-Image Benchmark.}
The LayouSAM-Eval benchmark serves as a comprehensive tool for evaluating L2I generation quality collected from a subset of LayouSAM. It comprises a total of 5,000 layout data.
We evaluate L2I generation quality using LayouSAM-Eval from two aspects:
\begin{itemize}
\item[--] \textit{Region-wise quality.} This aspect is evaluated for adherence to spatial and attribute accuracy via VLM’s~\cite{minicpm-v-2.6} Visual Question Answering (VQA). For each entity, spatially, the VLM evaluates whether the entity exists within the bounding box; for attributes, the VLM assesses whether the entity matches the color, text, and shape mentioned in the detailed descriptions.

\item[--] \textit{ Global-wise quality.} This aspect scores based on visual quality and global caption following, across multiple metrics including recently proposed scoring models like IR score~\cite{xu2023imagereward} and Pick score~\cite{kirstain2023pick}, as well as traditional metrics such as CLIP~\cite{radford2021clip}, FID~\cite{heusel2017fid} and IS~\cite{salimans2016is} scores.

\end{itemize}

\begin{figure}
  \centering
  \includegraphics[width=0.95\linewidth]{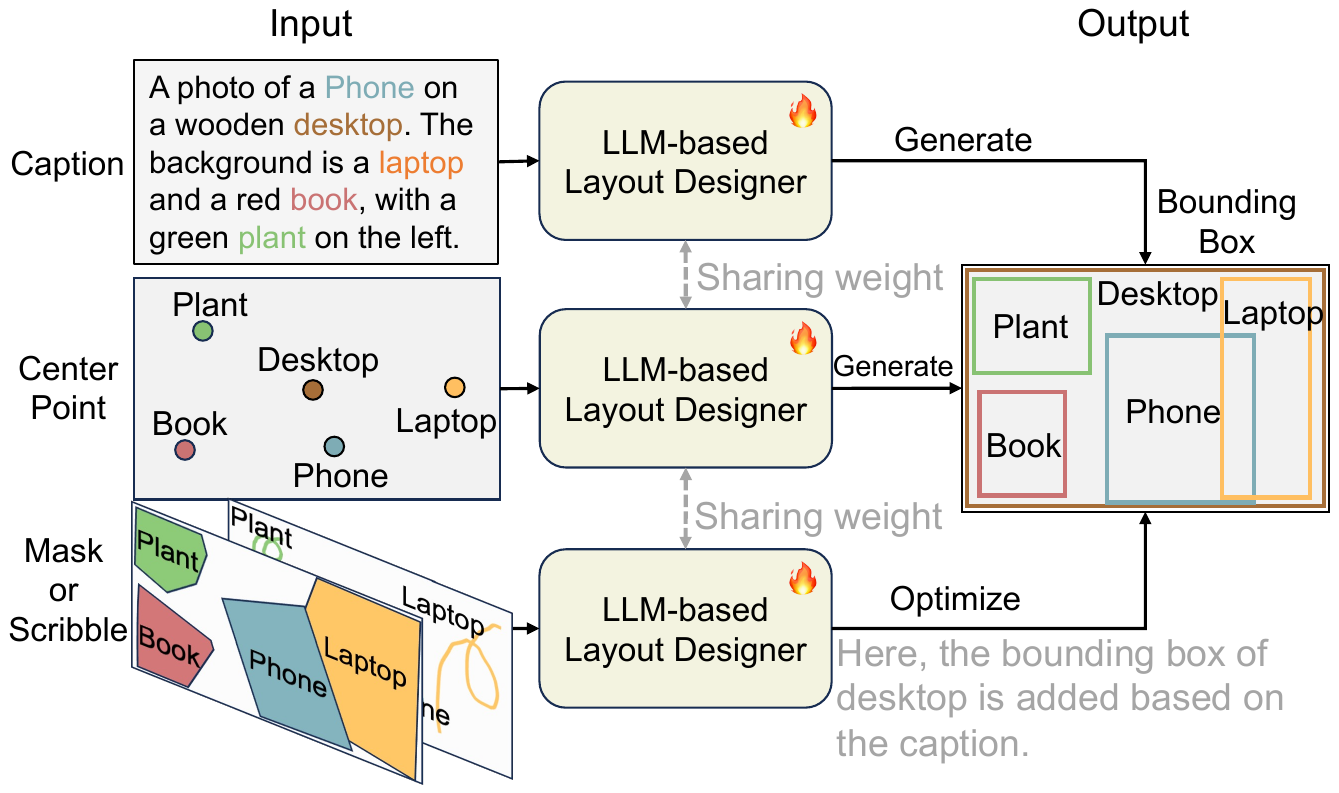}
  \caption{An overview of layout generation and optimization.}
  \label{fig:layout_desinger}
  \vspace{-1.8em}
\end{figure}

\subsection{Layout Designer}
\label{subsec:layout designer}
Recent studies~\cite{omost,feng2024ranni,yang2024rpg-master,feng2024layoutgpt,wu2024self-correct,chen2023textdiffuser2} have revealed that LLMs~\cite{achiam2023gpt4-turbo,dubey2024llama3.1-paper,team2024gemma-paper,hu2024minicpm-paper} exhibit expertise in layout planning due to their extensive training on multiple domains. Inspired by this, we further tame a LLM into a more comprehensive and professional layout designer $\mathrm{M}_l$.
As shown in \cref{fig:layout_desinger}, $\mathrm{M}_l$ is capable of executing two types of layout planning based on inputs of varying granularity:
\begin{itemize}
\item[--] \textit{Generation for coarse-grained inputs.} For user inputs with only a global caption, $\mathrm{M}_l$ designs from scratch based on the caption to determine which entities compose the layout and generates appropriate bounding boxes for these entities; for cases where the description and centroid coordinates~($x_0$,$y_0$) of each entity are provided, $\mathrm{M}_l$ designs harmonious bounding boxes based on this information.
\item[--] \textit{Optimization for fine-grained inputs.} For entities provided with detailed spatial information, such as masks and scribbles, we first transform them into bounding boxes according to predefined rules, then employ $\mathrm{M}_l$ to further optimize the layout based on the global caption and the descriptions of the entities, including additions, deletions, and modifications of the bounding boxes.
\end{itemize}
To enhance the expertise of $\mathrm{M}_l$, we construct 180,000 paired layout design data from LayoutSAM and fine-tune the pre-trained LLM~\cite{llama3.1} using LoRA~\cite{hu2021lora} with cross-entropy loss.

\section{Experiments}
\label{sec:experiments}

\subsection{Experimental Details}
\label{subsec:Experimental Details}

\paragraph{Layout-to-Image Generation.}
We conduct experiments on two types of datasets: the fine-grained open-set LayoutSAM and the coarse-grained closed-set COCO 2017~\cite{lin2014coco}.
For LayoutSAM, we train on 2.7 million image-text pairs with 10.7 million entities and conduct evaluations on LayoutSAM-Eval, which includes 5,000 prompts with detailed entity annotations. We measure generation quality using the metrics outlined in \cref{subsec:layout dataset}.
For COCO, following previous work~\cite{zheng2023layoutdiffusion,yang2023law-diffusion}, we filter out bounding boxes smaller than 2\% of the total image area and images containing dense bounding boxes and crowds, resulting in 61,002 training images and 2,565 validation images.
We use the YOLO-v11-x~\cite{khanam2024yolov11} to validate the model's layout adherence by detecting objects in the generated images and then calculating AP, $\mathrm{AP}^{50}$, and AR against the ground truth. 
We use IR~\cite{xu2023imagereward}, Pick~\cite{kirstain2023pick}, FID, CLIP, and IS to measure the global quality.

\vspace{-1.2em}
\paragraph{Text-to-Image Generation.} We conduct experiments on the T2I-CompBench~\cite{huang2023t2i-compbench} and evaluate image quality from five aspects: spatial, color, shape, texture, and numeracy.

\vspace{-1.2em}
\paragraph{Layout Generation and Optimization.} We construct 180,000 training sets based on the LayoutSAM training set for three types of user input and layout pairs: caption-layout pairs, center point-layout pairs, and suboptimal layout-layout pairs. Similarly, we construct 1,000 validation sets for each of these tasks from LayoutSAM-Eval.

\vspace{-1.2em}
\paragraph{Implementation Details.}
We employ experiments on SD3-medium and FLUX.1-dev. The training resolution for the LayoutSAM dataset is $1024 \times 1024$, and $512 \times 512$ for COCO. We utilize the AdamW optimizer with a fixed learning rate of 5e-5 and train the model for 600,000 iterations with a batch size of 16. We train SiamLayout with 8 A800-40G GPUs for 7 days. The value of $\lambda_{region}$ is set to 2. LayoutDesigner is fine-tuned on Llama-3.1-8B-Instruct~\cite{dubey2024llama3.1-paper} for one day using one A800-40G GPU.

\subsection{Evaluation on Layout-to-Image Generation}

\begin{table}[h]
\centering
\tabcolsep=0.06cm
\ra{1.1}
\scalebox{0.63}{
\begin{tabular}{@{}lccccccccc@{}}
\toprule
\multirow{2}{*}{LayoutSAM-Eval} & \multicolumn{4}{c}{Region-wise Quality} & \multicolumn{5}{c}{Global-wise Quality} \\ \cmidrule(l){2-5} \cmidrule(l){6-10} 
                    & Spatial~$\uparrow$   & Color~$\uparrow$   & Texture~$\uparrow$  & Shape~$\uparrow$  & IR~$\uparrow$     & Pick~$\uparrow$  & CLIP~$\uparrow$  & FID~$\downarrow$ & IS~$\uparrow$  \\ \midrule
\gray{Real Images}          &   \gray{98.95}	&\gray{98.45}	&\gray{98.90}	 &\gray{98.80}                  & -  &  -  & -  & -  & - \\  \midrule
GLIGEN~\cite{li2023gligen}             & 77.53     & 49.41   & 55.29    & 52.72  & -10.31   & 20.78  & \uuline{32.42} & 21.92 & \uuline{20.57} \\
Ranni~\cite{feng2024ranni}               & 41.38     & 24.10    & 25.57    & 23.35  & -28.46  & 20.49  & 31.40  & 27.24 & 19.81 \\
MIGC~\cite{zhou2024migc}                & 85.66     & 66.97   & 71.24    & 69.06  & -13.72  & 20.71  & 31.36 & 21.19 & 19.65 \\
InstanceDiff~\cite{wang2024instancediffusion}            & \uuline{87.99}     & 69.16   & \uuline{72.78}    & 71.08  & 9.14   & 21.01  & 31.40  & \uuline{19.67} & 20.02 \\
Be Yourself~\cite{dahary2024beyourself}         & 53.99     & 31.73   & 35.26    & 32.75  & -12.31  & 20.20   & 31.02 & 28.10  & 17.98 \\
HiCo~\cite{ma2025hico}           & 87.04     & \underline{69.19}   & 72.36    & \uuline{71.10}  & \uuline{12.36}   & \uuline{21.70}  & 32.18 & 22.61 & 20.15 \\\midrule
\textbf{SiamLayout-SD3}         & \underline{92.67}    & \underline{74.45}   & \underline{77.21}    & \underline{75.93}  & \underline{69.47}   & \underline{22.02}  & \textbf{34.01} & \underline{19.10}  & \textbf{22.04} \\
\textbf{SiamLayout-FLUX}         & \textbf{95.67}     & \textbf{80.71}   & \textbf{83.53}    & \textbf{82.80}  & \textbf{80.48}   & \textbf{22.16}  & \underline{33.92} & \textbf{16.12}  & \underline{21.81} \\
\textit{vs. prev. SoTA}      & \textbf{\green{+7.68}}      & \textbf{\green{+11.52}}    & \textbf{\green{+10.75}}     & \textbf{\green{+11.70}}   & \textbf{\green{+68.12}}    & \textbf{\green{+0.46}}   & \textbf{\green{+1.59}}  & \textbf{\green{+0.57}}  & \textbf{\green{+1.47}}  \\ \bottomrule
\end{tabular}}
\vspace{-0.5em}
\caption{Quantitative results on the LayoutSAM-Eval. \textbf{Bold}, \underline{underline}, and \uuline{double underline} represent the best, second, and third best methods, respectively.}
\label{tab: results on our benchmark}
\vspace{-1.5em}
\end{table}

\begin{figure}[h]
  \centering
  \includegraphics[width=1.0\linewidth]{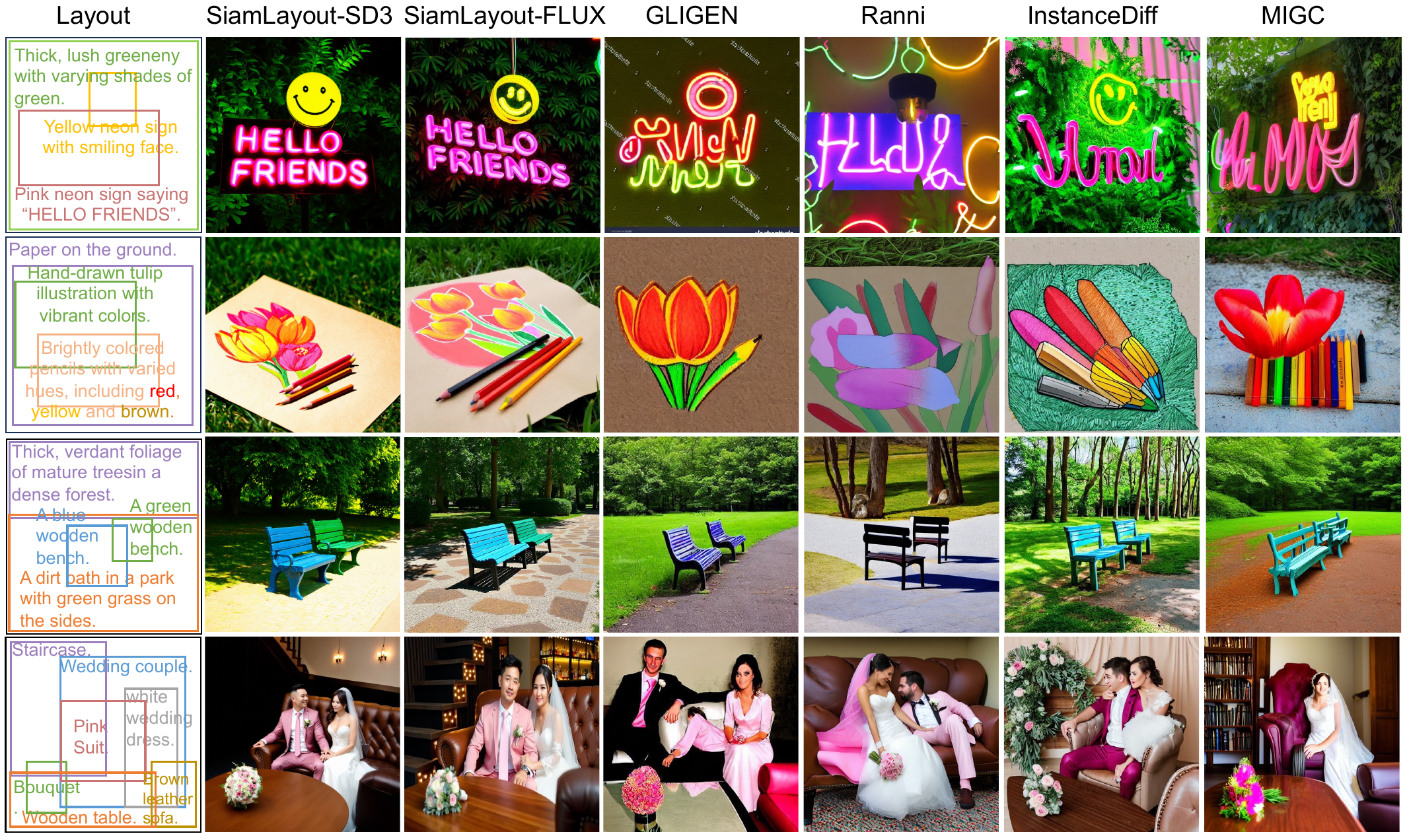}
  \caption{Qualitative results on the LayoutSAM-Eval. SiamLayout outperforms previous methods by a clear margin, especially in generating complex attributes such as color, texture, text, and portrait.}
  \label{fig:qualitative_results}
  \vspace{-1.0em}
\end{figure}

\paragraph{Fine-Grained Open-set L2I.}~\cref{tab: results on our benchmark} presents the quantitative results of SiamLayout on the fine-grained open-set LayoutSAM-Eval, including metrics of region-wise quality and global-wise quality. SiamLayout not only surpasses the current SOTA in terms of spatial response but also exhibits more precise responses in attributes such as color, texture, and shape. By fully unleashing the power of MM-DiT, SiamLayout also demonstrates a dominant advantage in overall image quality.
This is further confirmed by the qualitative results in \cref{fig:qualitative_results}, showing that SiamLayout achieves more accurate and aesthetically appealing attribute rendering in the regions localized by the bounding boxes, including the rendering of shapes, colors, textures, text, and portraits.

\vspace{-1.0em}
\paragraph{Coarse-Grained Closed-set L2I.}
We train and evaluate SiamLayout on COCO to confirm its generalization in coarse-grained closed-set layout-to-image generation, as shown in ~\cref{tab: results on coco}. In terms of image quality, SiamLayout outperforms previous methods by a clear margin on CLIP, FID, and IS metrics, thanks to the tailored framework that unleashes the capabilities of MM-DiT. In terms of spatial positioning response, SiamLayout is slightly inferior to InstanceDiff. We attribute this to two factors: \Rmnum{1}) The training dataset of InstanceDiff is a more fine-grained COCO dataset, with per-entity fine-grained attribute annotations; \Rmnum{2}) InstanceDiff generates each entity separately and then combines them, achieving more precise control at the cost of increased time and computational resources.

\begin{table}[h]
\centering
\tabcolsep=0.08cm
\ra{1.1}
\scalebox{0.83}{
\begin{tabular}{@{}lcccccc@{}}
\toprule
\multirow{2}{*}{COCO 2017}   & \multicolumn{3}{c}{Spatial} & \multicolumn{3}{c}{Global-wise Quality} \\ \cmidrule(l){2-4} \cmidrule(l){5-7} 
                  & AP~$\uparrow$      & $\mathrm{AP}^{50}$~$\uparrow$    & AR~$\uparrow$      & CLIP~$\uparrow$        & FID~$\downarrow$         & IS~$\uparrow$          \\ \midrule
GLIGEN~\cite{li2023gligen}            & 42.1   & 66.8   & 37.2   & \underline{30.25}       &  30.68       & 26.76       \\
MIGC~\cite{zhou2024migc}              & 39.1   & 66.0    & 36.8   & 28.60        & 31.43        & 26.14       \\
InstanceDiff~\cite{wang2024instancediffusion}& \textbf{51.9}   & \textbf{71.9}   & \textbf{43.7}   & 30.04       &   \underline{29.94}    & \underline{27.23}       \\ 
Ranni~\cite{feng2024ranni}             & 14.7   & 22.3   & 17.0    & 30.02       &  31.94     & 24.44       \\\midrule
\textbf{SiamLayout-SD3}              &    \underline{47.8}    &   \underline{69.8}      &   \underline{42.1}      & \textbf{33.32}       &  \textbf{29.27}      & \textbf{30.26}       \\ \bottomrule
\end{tabular}}
\vspace{-0.5em}
\caption{Evaluation on coarse-grained COCO benchmark}
\label{tab: results on coco}
\vspace{-1.0em}
\end{table}

\subsection{Evaluation on Text-to-Image Generation}
To further validate the impact of incorporating layout on text-to-image generation, we conduct experiments on the T2I-CompBench~\cite{huang2023t2i-compbench}. For the prompts, we first use LLM to plan the layout, which is then used to generate images via SiamLayout. ~\cref{tab: results on T2I-CompBench} reveals that, by introducing layout to provide further guidance signals for image generation, SD3 has seen a significant improvement in spatial adherence (from 32.00 to 47.36). Additionally, the benefits of layout are also reflected in the improved adherence to prompts regarding color, shape, texture, and the number of objects.

\vspace{-0.2em}
\begin{table}[h]
\centering
\tabcolsep=0.08cm
\ra{1.1}
\scalebox{0.82}{
\begin{tabular}{@{}lccccc@{}}
\toprule
T2I-CompBench  & Spatial~$\uparrow$ & Color~$\uparrow$ & Shape~$\uparrow$ & Texture~$\uparrow$ & Numeracy~$\uparrow$ \\ \midrule
Attn-Exct~\cite{chefer2023attn-exct}      & 14.55   & 64.00    & 45.17 & 59.63   & 47.67    \\
SDXL~\cite{podell2023sdxl}           & 21.33   & 58.79 & 46.87 & 52.99   & 49.88    \\
PixArt-$\alpha$~\cite{chen2023pixartalpha} & 20.64 & 66.90  & 49.27 & 64.77   & 50.58    \\
DALLE-3~\cite{betker2023dalle3}        & 28.65   & 77.85 & \textbf{62.05} & 70.36   & \underline{58.80}     \\ \midrule
SD3~\cite{esser2024sd3}            & \underline{32.00}    & \underline{81.32} & 58.85 & \underline{73.34}   & 58.34    \\
\textbf{SiamLayout-SD3}   & \textbf{47.36}   & \textbf{83.24} & \underline{61.32} & \textbf{75.51}   & \textbf{62.15}    \\ \bottomrule
\end{tabular}}
\vspace{-0.5em}
\caption{Quantitative results on T2I-CompBench.}
\label{tab: results on T2I-CompBench}
\vspace{-1.5em}
\end{table}

\subsection{Evaluation on Layout Planning}
\begin{figure}[h]
  \centering
  \includegraphics[width=1.0\linewidth]{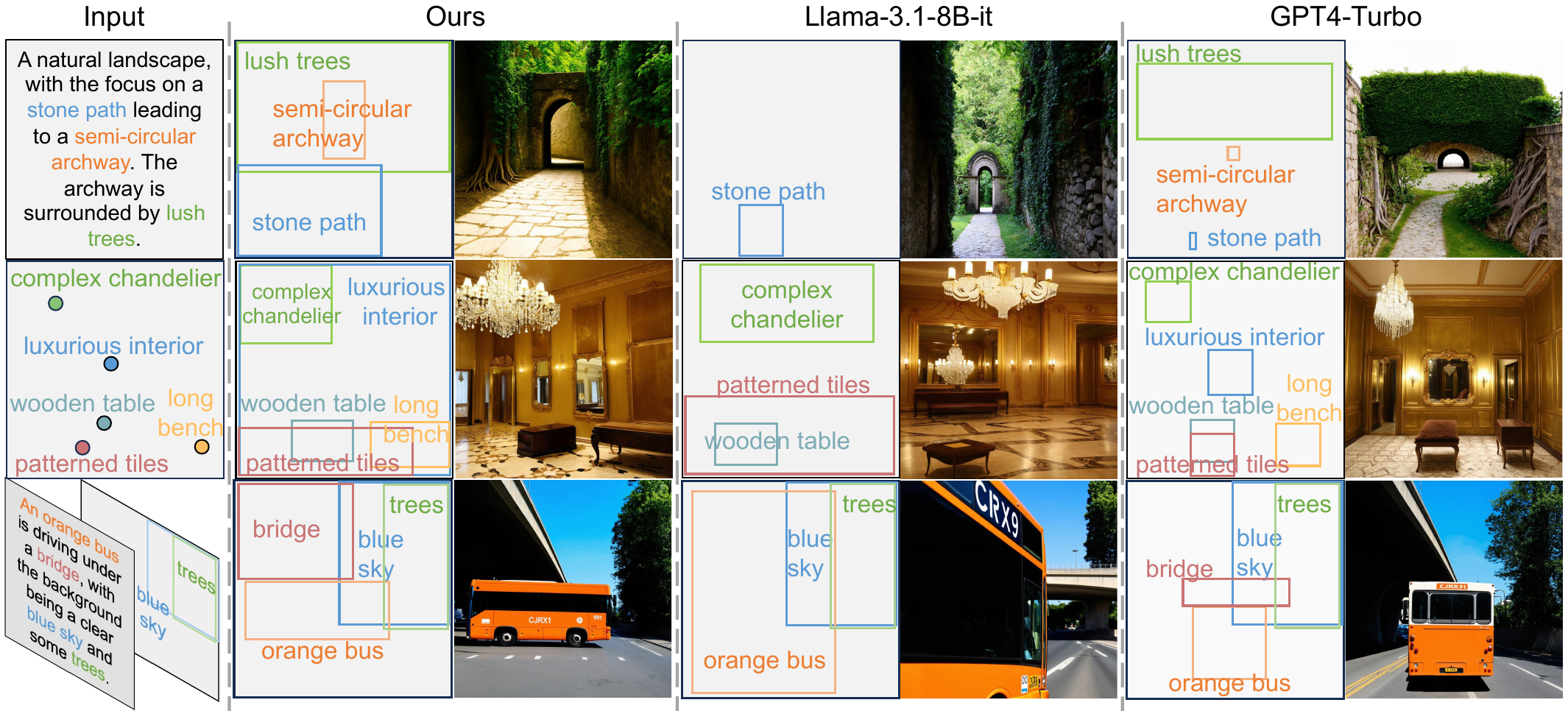}
  \vspace{-1.5em}
  \caption{Qualitative comparison of different layout planners.}
  \label{fig:auto_layout_visual}
  \vspace{-1.5em}
\end{figure}

To validate the capability of layout generation and optimization, we compare LayoutDesigner fine-tuned on Llama3.1 with the latest LLMs, as presented in \cref{tab: results of layout designer}.
Accuracy measures the correctness of the generated bounding boxes, including ensuring that the coordinates of the top-left corner are less than those of the bottom-right corner and that the bounding box does not exceed the image boundaries. Quality refers to the IR score of the image generated according to the planned layout, which is used to reflect the rationality and harmony of the layout.
In terms of format accuracy, LayoutDesigner shows significant improvement over the vanilla Llama and clearly outperforms the previous SOTA. Additionally, as the LayoutDesigner contributes more aesthetically pleasing layouts, the generated images possess higher quality. 
\cref{fig:auto_layout_visual} further confirms this, showing that layouts generated by Llama often fail to meet formatting standards or miss key elements, while those generated by GPT4-Turbo often violate fundamental physical laws (\eg overly small objects). 
In contrast, images generated from layouts designed by LayoutDesigner exhibit better quality as the layouts are more harmonious and aesthetically pleasing.

\begin{table}[h]
\centering
\tabcolsep=0.08cm
\ra{1.1}
\scalebox{0.75}{
\begin{tabular}{@{}lcccccc@{}}
\toprule
\multirow{2}{*}{Layout Planning} & \multicolumn{2}{c}{On global caption} & \multicolumn{2}{c}{On center point} & \multicolumn{2}{c}{On bounding box} \\\cmidrule(l){2-3} \cmidrule(l){4-5} \cmidrule(l){6-7}
                                   & Acc~$\uparrow$  & Quality~$\uparrow$  & Acc~$\uparrow$   & Quality~$\uparrow$  & Acc~$\uparrow$ & Quality~$\uparrow$            \\\midrule 
Gemma-7b-it~\cite{gemma-7b-it}        & 64.79          & 46.33            & 96.14             & 53.46              & 98.83             & 65.10              \\
Qwen2.5-7b-it~\cite{qwen2.5-7b-it}    & 45.09          & 53.49            & 75.36             & 50.63              & 96.50             & 63.22              \\
MiniCPM3-4b~\cite{minicpm-v-2.6}      & 70.71          & 55.80            & 66.72             & \underline{59.24}              & 83.25             & 61.57              \\
Llama3.1-8b-it~\cite{llama3.1}        & 50.11          & 55.09            & 73.35             & 50.22              & 97.68             & \underline{65.40}              \\ 
GPT4-Turbo~\cite{achiam2023gpt4-turbo} & \underline{76.60}& \underline{60.03} & \underline{99.53}  & 57.41   & \underline{99.96}   & 65.34    \\\midrule
\textbf{LayoutDesigner}               & \textbf{100.0} & \textbf{66.46}   & \textbf{100.0}   & \textbf{65.29}       & \textbf{100.0}       &\textbf{68.52}       \\
\textit{vs. prev. SoTA}               &  \textbf{\green{+23.40}}   &  \textbf{\green{+6.43}}   & \textbf{\green{+0.47}} & \textbf{\green{+6.05}}& \textbf{\green{+0.04}}  &\textbf{\green{+3.12}}  \\
\textit{vs. w/o. SFT} &  \textbf{\green{+49.89}}   &  \textbf{\green{+11.37}}   & \textbf{\green{+26.65}} & \textbf{\green{+15.07}}& \textbf{\green{+2.32}}  &\textbf{\green{+3.12}}   \\\bottomrule
\end{tabular}}
\vspace{-0.5em}
\caption{Quantitative comparison of different layout planners.}
\label{tab: results of layout designer}
\vspace{-1.0em}
\end{table}

\subsection{Ablation Study}

\begin{figure}[h]
  \centering
  \includegraphics[width=0.98\linewidth]{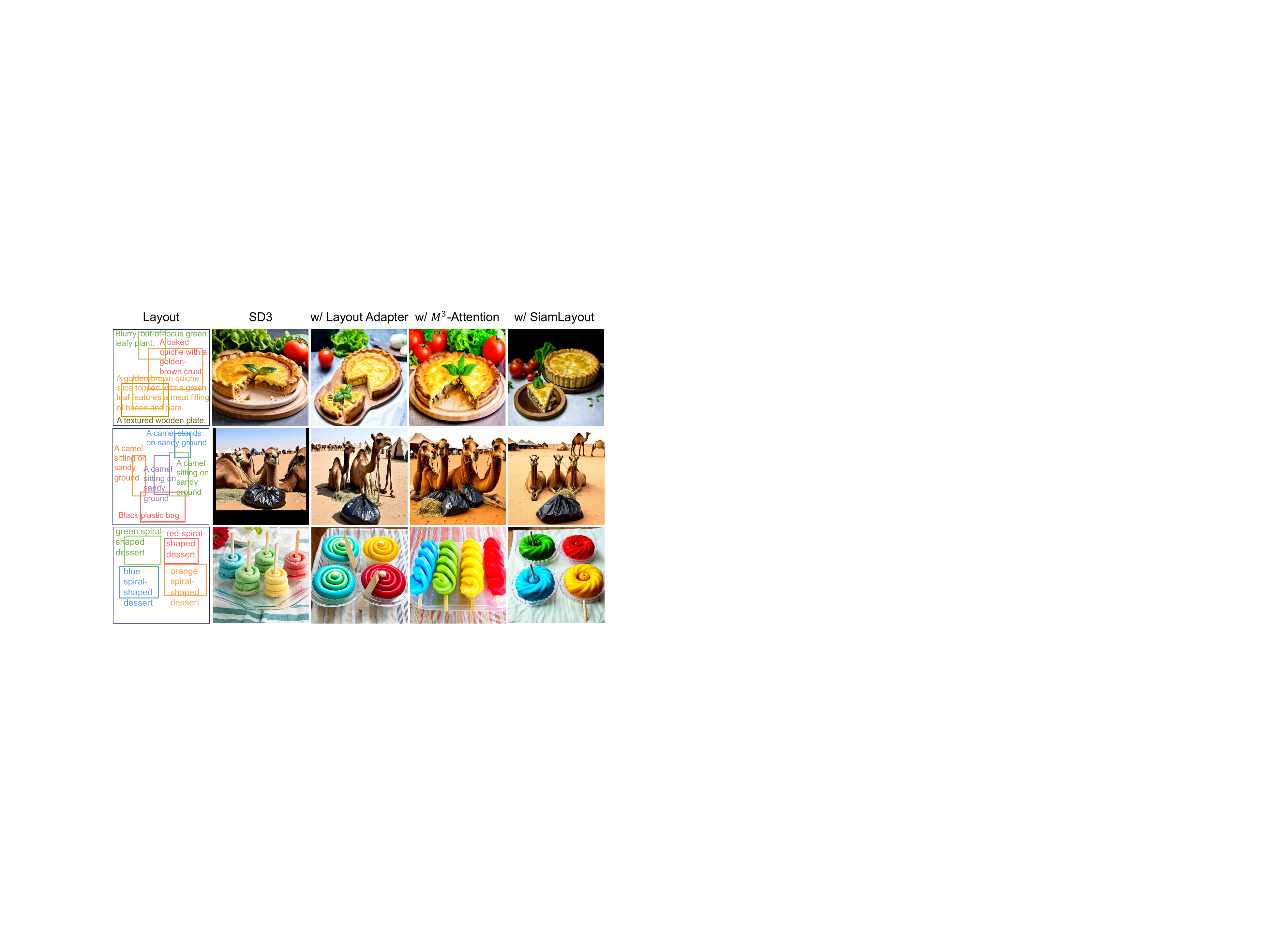}
  \vspace{-0.5em}
  \caption{Qualitative results of three network variants.}
  \label{fig:qualitative_results_on_module}
  \vspace{-1.5em}
\end{figure}

\paragraph{Ablations on Network Design.}
We explore three network variants aimed at integrating layout guidance into MM-DiT. \cref{tab: quantitative_results_on_module} illustrates the accuracy of layout adherence of these designs. Compared to the vanilla SD3, Layout Adapter enhances the model’s adherence to spatial locations and attributes by introducing cross-attention between the image and layout. However, as shown in \cref{fig:qualitative_results_on_module}, it falls short when dealing with complex color, quantity, and texture requirements. We attribute this to the fact that the layout is not considered an independent modality equally important as the global caption and image, which limits the potential of layout guidance.
The initial intention behind designing $\mathrm{M}^3$-Attention is to make layout, image, and caption play an equal role.
However, due to the competition among these modalities in the attention map of $\mathrm{M}^3$-Attention, layout modality consistently is at a disadvantage, which is reflected in both quantitative and qualitative results as lower layout responsiveness.
SiamLayout decouples $\mathrm{M}^3$-Attention into two parallel MM-Attention branches: image-text and image-layout. This design allows each branch to play an independent role without interfering with each other, jointly contributing to precise responses to spatial locations and complex attributes.

\begin{table}[h]
\vspace{-0.5em}
\centering
\tabcolsep=0.08cm
\ra{1.1}
\scalebox{0.9}{
\begin{tabular}{@{}lcccc@{}}
\toprule
                        & Spatial~$\uparrow$       & Color~$\uparrow$         & Texture~$\uparrow$               &Shape~$\uparrow$ \\\midrule
Stable Diffusion 3      & 78.70                     & 59.22                     & 60.66                       & 58.73                \\\midrule
w/  Layout Adapter          & \underline{88.43}       & \underline{71.67}         & \underline{73.56}         & \underline{72.61}          \\
w/  $\mathrm{M}^3$-Attention            & 79.15              & 60.19                     & 62.96                       & 61.29         \\
\rowcolor[HTML]{e9e9e9} 
\textbf{w/  SiamLayout}    & \textbf{92.67}    & \textbf{74.45}              & \textbf{77.21}         & \textbf{75.93}         \\\bottomrule
\end{tabular}}
\vspace{-0.3em}
\caption{Ablation study on different network variants.}
\label{tab: quantitative_results_on_module}
\vspace{-1.5em}
\end{table}

\begin{figure}[h]
  \centering
  \vspace{-0.5em}
  \includegraphics[width=0.92\linewidth]{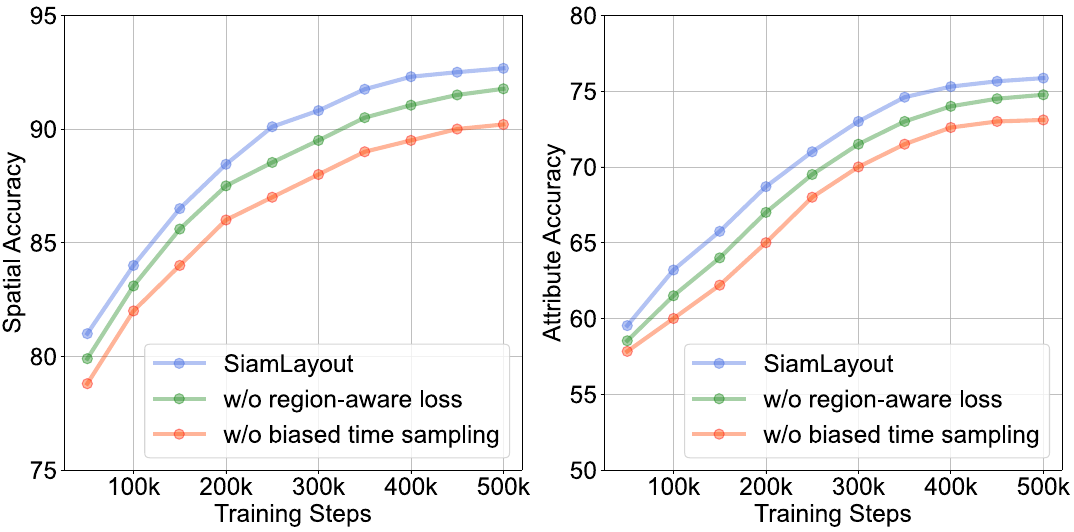}
  \vspace{-0.5em}
  \caption{Ablation study on different training strategies.}
  \label{fig:two_strategies_ablation}
  \vspace{-1.5em}
\end{figure}

\vspace{-1.0em}
\paragraph{The Impact of Training Strategies.}
In \cref{fig:two_strategies_ablation}, we explore the impact of two strategies introduced during training on SiamLayout: biased time step sampling and region-aware loss. With the region-aware loss $\mathcal{L}_{region}$, the model focuses more on the areas localized by the layout, accelerating the model's convergence.
In addition, since layout mainly guides structural content generated at larger time steps, sampling these steps with higher probability (\ie biased time step sampling) also speeds up convergence.

\section{Conclusion}
\label{sec:conclusion}
We presented \textbf{CreatiLayout}, a layout-to-image (L2I) system that integrates three key components: a layout model (SiamLayout), a dataset (LayoutSAM), and a planner (LayoutDesigner). SiamLayout treats layout as an independent modality and guides image through an image-layout branch that is siamese to the image-text branch, unleashing the power of MM-DiT for high-quality and accurate L2I generation. It outperforms previous work in generating complex attributes such as color, shape, and text. LayoutSAM is a large-scale layout dataset with 2.7M image-text pairs and 10.7M entities, accompanied by LayoutSAM-Eval for comprehensive evaluation of quality. LayoutDesigner leverages a large language model as a professional layout planner, capable of handling diverse user inputs with varying granularity.

\vspace{-1.2em}
\paragraph{Limitation and Future Work.}
Introducing a large language model for layout planning brings extra computation costs. Integrating layout planning with L2I generation into an end-to-end model is an important direction for future research. 
In addition, improving the accuracy of control in layouts with a large number of entities, or in scenarios where entities contain rich text, is also a promising direction.

\paragraph{Acknowledgements.} This work was supported in part by the National Natural Science Foundation of China (Grant 62032006 and Grant 62472098) and ByteDance (No.CT20230914000147).

{
    \small
    \bibliographystyle{ieeenat_fullname}
    \bibliography{main}

\begin{thebibliography}{81}
\providecommand{\natexlab}[1]{#1}
\providecommand{\url}[1]{\texttt{#1}}
\expandafter\ifx\csname urlstyle\endcsname\relax
  \providecommand{\doi}[1]{doi: #1}\else
  \providecommand{\doi}{doi: \begingroup \urlstyle{rm}\Url}\fi

\bibitem[Achiam et~al.(2023)Achiam, Adler, Agarwal, Ahmad, Akkaya, Aleman, Almeida, Altenschmidt, Altman, Anadkat, et~al.]{achiam2023gpt4-turbo}
Josh Achiam, Steven Adler, Sandhini Agarwal, Lama Ahmad, Ilge Akkaya, Florencia~Leoni Aleman, Diogo Almeida, Janko Altenschmidt, Sam Altman, Shyamal Anadkat, et~al.
\newblock Gpt-4 technical report.
\newblock \emph{arXiv preprint arXiv:2303.08774}, 2023.

\bibitem[Bao et~al.(2023)Bao, Nie, Xue, Cao, Li, Su, and Zhu]{bao2023u-vit}
Fan Bao, Shen Nie, Kaiwen Xue, Yue Cao, Chongxuan Li, Hang Su, and Jun Zhu.
\newblock All are worth words: A vit backbone for diffusion models.
\newblock In \emph{CVPR}, 2023.

\bibitem[Betker et~al.(2023)Betker, Goh, Jing, Brooks, Wang, Li, Ouyang, Zhuang, Lee, Guo, et~al.]{betker2023dalle3}
James Betker, Gabriel Goh, Li Jing, Tim Brooks, Jianfeng Wang, Linjie Li, Long Ouyang, Juntang Zhuang, Joyce Lee, Yufei Guo, et~al.
\newblock Improving image generation with better captions.
\newblock \emph{Computer Science. https://cdn. openai. com/papers/dall-e-3. pdf}, 2023.

\bibitem[Chefer et~al.(2023)Chefer, Alaluf, Vinker, Wolf, and Cohen-Or]{chefer2023attn-exct}
Hila Chefer, Yuval Alaluf, Yael Vinker, Lior Wolf, and Daniel Cohen-Or.
\newblock Attend-and-excite: Attention-based semantic guidance for text-to-image diffusion models.
\newblock \emph{ACM Transactions on Graphics (TOG)}, 2023.

\bibitem[Chen et~al.(2024{\natexlab{a}})Chen, Huang, Lv, Cui, Chen, and Wei]{chen2023textdiffuser2}
Jingye Chen, Yupan Huang, Tengchao Lv, Lei Cui, Qifeng Chen, and Furu Wei.
\newblock Textdiffuser-2: Unleashing the power of language models for text rendering.
\newblock In \emph{ECCV}, 2024{\natexlab{a}}.

\bibitem[Chen et~al.(2024{\natexlab{b}})Chen, Jincheng, Chongjian, Yao, Xie, Wang, Kwok, Luo, Lu, and Li]{chen2023pixartalpha}
Junsong Chen, YU Jincheng, GE Chongjian, Lewei Yao, Enze Xie, Zhongdao Wang, James Kwok, Ping Luo, Huchuan Lu, and Zhenguo Li.
\newblock Pixart-$\alpha$: Fast training of diffusion transformer for photorealistic text-to-image synthesis.
\newblock In \emph{ICLR}, 2024{\natexlab{b}}.

\bibitem[Chen et~al.(2023)Chen, Liu, Yang, Yuan, You, Liu, and Yang]{chen2023reasonyourlayout}
Xiaohui Chen, Yongfei Liu, Yingxiang Yang, Jianbo Yuan, Quanzeng You, Li-Ping Liu, and Hongxia Yang.
\newblock Reason out your layout: Evoking the layout master from large language models for text-to-image synthesis.
\newblock \emph{arXiv preprint arXiv:2311.17126}, 2023.

\bibitem[Cho et~al.(2023)Cho, Zala, and Bansal]{cho2024visualprogramming}
Jaemin Cho, Abhay Zala, and Mohit Bansal.
\newblock Visual programming for step-by-step text-to-image generation and evaluation.
\newblock In \emph{NeurIPS}, 2023.

\bibitem[Dahary et~al.(2024)Dahary, Patashnik, Aberman, and Cohen-Or]{dahary2024beyourself}
Omer Dahary, Or Patashnik, Kfir Aberman, and Daniel Cohen-Or.
\newblock Be yourself: Bounded attention for multi-subject text-to-image generation.
\newblock In \emph{ECCV}, 2024.

\bibitem[Dubey et~al.(2024)Dubey, Jauhri, Pandey, Kadian, Al-Dahle, Letman, Mathur, Schelten, Yang, Fan, et~al.]{dubey2024llama3.1-paper}
Abhimanyu Dubey, Abhinav Jauhri, Abhinav Pandey, Abhishek Kadian, Ahmad Al-Dahle, Aiesha Letman, Akhil Mathur, Alan Schelten, Amy Yang, Angela Fan, et~al.
\newblock The llama 3 herd of models.
\newblock \emph{arXiv preprint arXiv:2407.21783}, 2024.

\bibitem[Esser et~al.(2024)Esser, Kulal, Blattmann, Entezari, M{\"u}ller, Saini, Levi, Lorenz, Sauer, Boesel, et~al.]{esser2024sd3}
Patrick Esser, Sumith Kulal, Andreas Blattmann, Rahim Entezari, Jonas M{\"u}ller, Harry Saini, Yam Levi, Dominik Lorenz, Axel Sauer, Frederic Boesel, et~al.
\newblock Scaling rectified flow transformers for high-resolution image synthesis.
\newblock \emph{arXiv preprint arXiv:2403.03206}, 2024.

\bibitem[Feng et~al.(2023)Feng, Zhu, Fu, Jampani, Akula, He, Basu, Wang, and Wang]{feng2024layoutgpt}
Weixi Feng, Wanrong Zhu, Tsu-jui Fu, Varun Jampani, Arjun Akula, Xuehai He, Sugato Basu, Xin~Eric Wang, and William~Yang Wang.
\newblock Layoutgpt: Compositional visual planning and generation with large language models.
\newblock In \emph{NeurIPS}, 2023.

\bibitem[Feng et~al.(2024)Feng, Gong, Chen, Shen, Liu, and Zhou]{feng2024ranni}
Yutong Feng, Biao Gong, Di Chen, Yujun Shen, Yu Liu, and Jingren Zhou.
\newblock Ranni: Taming text-to-image diffusion for accurate instruction following.
\newblock In \emph{CVPR}, 2024.

\bibitem[Gao et~al.(2024)Gao, Zhuo, Lin, Liu, Chen, Du, Xie, Luo, Qiu, Zhang, et~al.]{gao2024lumina}
Peng Gao, Le Zhuo, Ziyi Lin, Chris Liu, Junsong Chen, Ruoyi Du, Enze Xie, Xu Luo, Longtian Qiu, Yuhang Zhang, et~al.
\newblock Lumina-t2x: Transforming text into any modality, resolution, and duration via flow-based large diffusion transformers.
\newblock \emph{arXiv preprint arXiv:2405.05945}, 2024.

\bibitem[Gong et~al.(2024)Gong, Huang, Feng, Zhang, Li, and Liu]{gong2024check}
Biao Gong, Siteng Huang, Yutong Feng, Shiwei Zhang, Yuyuan Li, and Yu Liu.
\newblock Check locate rectify: A training-free layout calibration system for text-to-image generation.
\newblock In \emph{CVPR}, 2024.

\bibitem[Google(2024)]{gemma-7b-it}
Google.
\newblock Gemma-7b-it.
\newblock \url{https://huggingface.co/google/gemma-7b-it}, 2024.

\bibitem[Group(2024{\natexlab{a}})]{Qwen-VL-Captioner}
Alibaba Group.
\newblock Qwen-vl-chat-finetuned-dense-captioner.
\newblock \url{https://modelscope.cn/models/Tongyi-DataEngine/Qwen-VL-Chat-Finetuned-Dense-Captioner}, 2024{\natexlab{a}}.

\bibitem[Group(2024{\natexlab{b}})]{qwen2.5-7b-it}
Alibaba Group.
\newblock Qwen2.5-7b-it.
\newblock \url{https://huggingface.co/Qwen/Qwen2.5-7B-Instruct}, 2024{\natexlab{b}}.

\bibitem[Guerreiro et~al.(2024)Guerreiro, Inoue, Masui, Otani, and Nakayama]{guerreiro2025layoutflow}
Julian Jorge~Andrade Guerreiro, Naoto Inoue, Kento Masui, Mayu Otani, and Hideki Nakayama.
\newblock Layoutflow: Flow matching for layout generation.
\newblock In \emph{ECCV}, 2024.

\bibitem[Heusel et~al.(2017)Heusel, Ramsauer, Unterthiner, Nessler, and Hochreiter]{heusel2017fid}
Martin Heusel, Hubert Ramsauer, Thomas Unterthiner, Bernhard Nessler, and Sepp Hochreiter.
\newblock Gans trained by a two time-scale update rule converge to a local nash equilibrium.
\newblock In \emph{NeurIPS}, 2017.

\bibitem[Ho et~al.(2020)Ho, Jain, and Abbeel]{ho2020ddpm}
Jonathan Ho, Ajay Jain, and Pieter Abbeel.
\newblock Denoising diffusion probabilistic models.
\newblock In \emph{NeurIPS}, 2020.

\bibitem[Hoe et~al.(2024)Hoe, Jiang, Chan, Tan, and Hu]{hoe2024interactdiffusion}
Jiun~Tian Hoe, Xudong Jiang, Chee~Seng Chan, Yap-Peng Tan, and Weipeng Hu.
\newblock Interactdiffusion: Interaction control in text-to-image diffusion models.
\newblock In \emph{CVPR}, 2024.

\bibitem[Horita et~al.(2024)Horita, Inoue, Kikuchi, Yamaguchi, and Aizawa]{horita2024retrieval-augmented}
Daichi Horita, Naoto Inoue, Kotaro Kikuchi, Kota Yamaguchi, and Kiyoharu Aizawa.
\newblock Retrieval-augmented layout transformer for content-aware layout generation.
\newblock In \emph{CVPR}, 2024.

\bibitem[Hu et~al.(2021)Hu, Shen, Wallis, Allen-Zhu, Li, Wang, Wang, and Chen]{hu2021lora}
Edward~J Hu, Yelong Shen, Phillip Wallis, Zeyuan Allen-Zhu, Yuanzhi Li, Shean Wang, Lu Wang, and Weizhu Chen.
\newblock Lora: Low-rank adaptation of large language models.
\newblock \emph{arXiv preprint arXiv:2106.09685}, 2021.

\bibitem[Hu et~al.(2024)Hu, Tu, Han, He, Cui, Long, Zheng, Fang, Huang, Zhao, et~al.]{hu2024minicpm-paper}
Shengding Hu, Yuge Tu, Xu Han, Chaoqun He, Ganqu Cui, Xiang Long, Zhi Zheng, Yewei Fang, Yuxiang Huang, Weilin Zhao, et~al.
\newblock Minicpm: Unveiling the potential of small language models with scalable training strategies.
\newblock \emph{arXiv preprint arXiv:2404.06395}, 2024.

\bibitem[Huang et~al.(2023)Huang, Sun, Xie, Li, and Liu]{huang2023t2i-compbench}
Kaiyi Huang, Kaiyue Sun, Enze Xie, Zhenguo Li, and Xihui Liu.
\newblock T2i-compbench: A comprehensive benchmark for open-world compositional text-to-image generation.
\newblock In \emph{NeurIPS}, 2023.

\bibitem[Hui et~al.(2024)Hui, Yang, Cui, Yang, Liu, Zhang, Liu, Zhang, Yu, Dang, et~al.]{hui2024qwen2.5-paper}
Binyuan Hui, Jian Yang, Zeyu Cui, Jiaxi Yang, Dayiheng Liu, Lei Zhang, Tianyu Liu, Jiajun Zhang, Bowen Yu, Kai Dang, et~al.
\newblock Qwen2. 5-coder technical report.
\newblock \emph{arXiv preprint arXiv:2409.12186}, 2024.

\bibitem[Jia et~al.(2024)Jia, Luo, Dang, Dai, Chang, Wang, and Wang]{jia2024ssmg}
Chengyou Jia, Minnan Luo, Zhuohang Dang, Guang Dai, Xiaojun Chang, Mengmeng Wang, and Jingdong Wang.
\newblock Ssmg: Spatial-semantic map guided diffusion model for free-form layout-to-image generation.
\newblock In \emph{AAAI}, 2024.

\bibitem[Khanam and Hussain(2024)]{khanam2024yolov11}
Rahima Khanam and Muhammad Hussain.
\newblock Yolov11: An overview of the key architectural enhancements.
\newblock \emph{arXiv preprint arXiv:2410.17725}, 2024.

\bibitem[Kingma(2013)]{kingma2013vae}
Diederik~P Kingma.
\newblock Auto-encoding variational bayes.
\newblock \emph{arXiv preprint arXiv:1312.6114}, 2013.

\bibitem[Kirillov et~al.(2023)Kirillov, Mintun, Ravi, Mao, Rolland, Gustafson, Xiao, Whitehead, Berg, Lo, et~al.]{kirillov2023sam}
Alexander Kirillov, Eric Mintun, Nikhila Ravi, Hanzi Mao, Chloe Rolland, Laura Gustafson, Tete Xiao, Spencer Whitehead, Alexander~C Berg, Wan-Yen Lo, et~al.
\newblock Segment anything.
\newblock In \emph{ICCV}, 2023.

\bibitem[Kirstain et~al.(2023)Kirstain, Polyak, Singer, Matiana, Penna, and Levy]{kirstain2023pick}
Yuval Kirstain, Adam Polyak, Uriel Singer, Shahbuland Matiana, Joe Penna, and Omer Levy.
\newblock Pick-a-pic: An open dataset of user preferences for text-to-image generation.
\newblock \emph{arXiv preprint arXiv:2305.01569}, 2023.

\bibitem[Labs(2024)]{flux}
Black~Forest Labs.
\newblock Flux.
\newblock \url{https://blackforestlabs.ai/announcing-black-forest-labs}, 2024.

\bibitem[LAION(2022)]{LAION-Aesthetics}
LAION.
\newblock Laion-aesthetics v2.
\newblock \url{https://github.com/christophschuhmann/improved-aesthetic-predictor}, 2022.

\bibitem[Lee et~al.(2024)Lee, Yoon, and Sung]{lee2024groundit}
Phillip~Y Lee, Taehoon Yoon, and Minhyuk Sung.
\newblock Groundit: Grounding diffusion transformers via noisy patch transplantation.
\newblock \emph{arXiv preprint arXiv:2410.20474}, 2024.

\bibitem[Li et~al.(2024{\natexlab{a}})Li, Kamko, Akhgari, Sabet, Xu, and Doshi]{li2024playground}
Daiqing Li, Aleks Kamko, Ehsan Akhgari, Ali Sabet, Linmiao Xu, and Suhail Doshi.
\newblock Playground v2. 5: Three insights towards enhancing aesthetic quality in text-to-image generation.
\newblock \emph{arXiv preprint arXiv:2402.17245}, 2024{\natexlab{a}}.

\bibitem[Li et~al.(2025)Li, Xing, Wang, Zhang, Dai, and Wu]{li2025magicmotion}
Quanhao Li, Zhen Xing, Rui Wang, Hui Zhang, Qi Dai, and Zuxuan Wu.
\newblock Magicmotion: Controllable video generation with dense-to-sparse trajectory guidance.
\newblock \emph{arXiv preprint arXiv:2503.16421}, 2025.

\bibitem[Li et~al.(2017)Li, Wang, Li, Agustsson, and Van~Gool]{li2017webvision}
Wen Li, Limin Wang, Wei Li, Eirikur Agustsson, and Luc Van~Gool.
\newblock Webvision database: Visual learning and understanding from web data.
\newblock \emph{arXiv preprint arXiv:1708.02862}, 2017.

\bibitem[Li et~al.(2023)Li, Liu, Wu, Mu, Yang, Gao, Li, and Lee]{li2023gligen}
Yuheng Li, Haotian Liu, Qingyang Wu, Fangzhou Mu, Jianwei Yang, Jianfeng Gao, Chunyuan Li, and Yong~Jae Lee.
\newblock Gligen: Open-set grounded text-to-image generation.
\newblock In \emph{CVPR}, 2023.

\bibitem[Li et~al.(2024{\natexlab{b}})Li, Zhang, Lin, Xiong, Long, Deng, Zhang, Liu, Huang, Xiao, et~al.]{li2024hunyuandit}
Zhimin Li, Jianwei Zhang, Qin Lin, Jiangfeng Xiong, Yanxin Long, Xinchi Deng, Yingfang Zhang, Xingchao Liu, Minbin Huang, Zedong Xiao, et~al.
\newblock Hunyuan-dit: A powerful multi-resolution diffusion transformer with fine-grained chinese understanding.
\newblock \emph{arXiv e-prints}, 2024{\natexlab{b}}.

\bibitem[Lian et~al.(2024)Lian, Li, Yala, and Darrell]{lianllm-grounded}
Long Lian, Boyi Li, Adam Yala, and Trevor Darrell.
\newblock Llm-grounded diffusion: Enhancing prompt understanding of text-to-image diffusion models with large language models.
\newblock \emph{TMLR}, 2024.

\bibitem[Lin et~al.(2014)Lin, Maire, Belongie, Hays, Perona, Ramanan, Doll{\'a}r, and Zitnick]{lin2014coco}
Tsung-Yi Lin, Michael Maire, Serge Belongie, James Hays, Pietro Perona, Deva Ramanan, Piotr Doll{\'a}r, and C~Lawrence Zitnick.
\newblock Microsoft coco: Common objects in context.
\newblock In \emph{ECCV}, 2014.

\bibitem[Liu et~al.(2024)Liu, Akhgari, Visheratin, Kamko, Xu, Shrirao, Souza, Doshi, and Li]{liu2024playgroundv3}
Bingchen Liu, Ehsan Akhgari, Alexander Visheratin, Aleks Kamko, Linmiao Xu, Shivam Shrirao, Joao Souza, Suhail Doshi, and Daiqing Li.
\newblock Playground v3: Improving text-to-image alignment with deep-fusion large language models.
\newblock \emph{arXiv preprint arXiv:2409.10695}, 2024.

\bibitem[Liu et~al.(2023)Liu, Zeng, Ren, Li, Zhang, Yang, Jiang, Li, Yang, Su, et~al.]{liu2023grounding1}
Shilong Liu, Zhaoyang Zeng, Tianhe Ren, Feng Li, Hao Zhang, Jie Yang, Qing Jiang, Chunyuan Li, Jianwei Yang, Hang Su, et~al.
\newblock Grounding dino: Marrying dino with grounded pre-training for open-set object detection.
\newblock \emph{arXiv preprint arXiv:2303.05499}, 2023.

\bibitem[Ma et~al.(2025)Ma, Liu, Ma, Wu, Leng, and Yin]{ma2025hico}
Yuhang Ma, Shanyuan Liu, Ao Ma, Xiaoyu Wu, Dawei Leng, and Yuhui Yin.
\newblock Hico: Hierarchical controllable diffusion model for layout-to-image generation.
\newblock In \emph{NeurIPS}, 2025.

\bibitem[Meta(2024)]{llama3.1}
Meta.
\newblock Llama-3.1-8b-instruct.
\newblock \url{https://huggingface.co/meta-llama/Llama-3.1-8B-Instruct}, 2024.

\bibitem[Mildenhall et~al.(2021)Mildenhall, Srinivasan, Tancik, Barron, Ramamoorthi, and Ng]{mildenhall2021fourier}
Ben Mildenhall, Pratul~P Srinivasan, Matthew Tancik, Jonathan~T Barron, Ravi Ramamoorthi, and Ren Ng.
\newblock Nerf: Representing scenes as neural radiance fields for view synthesis.
\newblock \emph{Communications of the ACM}, 2021.

\bibitem[Nie et~al.(2024)Nie, Liu, Mardani, Liu, Eckart, and Vahdat]{nie2024blobgen}
Weili Nie, Sifei Liu, Morteza Mardani, Chao Liu, Benjamin Eckart, and Arash Vahdat.
\newblock Compositional text-to-image generation with dense blob representations.
\newblock \emph{arXiv preprint arXiv:2405.08246}, 2024.

\bibitem[OpenBMB(2024)]{minicpm-v-2.6}
OpenBMB.
\newblock Minicpm-v-2.6.
\newblock \url{https://huggingface.co/openbmb/MiniCPM-V-2_6}, 2024.

\bibitem[Peebles and Xie(2023)]{peebles2023dit}
William Peebles and Saining Xie.
\newblock Scalable diffusion models with transformers.
\newblock In \emph{CVPR}, 2023.

\bibitem[Phung et~al.(2024)Phung, Ge, and Huang]{phung2024attentionrefocusing}
Quynh Phung, Songwei Ge, and Jia-Bin Huang.
\newblock Grounded text-to-image synthesis with attention refocusing.
\newblock In \emph{CVPR}, 2024.

\bibitem[Podell et~al.(2024)Podell, English, Lacey, Blattmann, Dockhorn, M{\"u}ller, Penna, and Rombach]{podell2023sdxl}
Dustin Podell, Zion English, Kyle Lacey, Andreas Blattmann, Tim Dockhorn, Jonas M{\"u}ller, Joe Penna, and Robin Rombach.
\newblock Sdxl: Improving latent diffusion models for high-resolution image synthesis.
\newblock In \emph{ICLR}, 2024.

\bibitem[Qu et~al.(2023)Qu, Wu, Fei, Nie, and Chua]{qu2023layoutllm}
Leigang Qu, Shengqiong Wu, Hao Fei, Liqiang Nie, and Tat-Seng Chua.
\newblock Layoutllm-t2i: Eliciting layout guidance from llm for text-to-image generation.
\newblock In \emph{ACM MM}, 2023.

\bibitem[Radford et~al.(2021)Radford, Kim, Hallacy, Ramesh, Goh, Agarwal, Sastry, Askell, Mishkin, Clark, et~al.]{radford2021clip}
Alec Radford, Jong~Wook Kim, Chris Hallacy, Aditya Ramesh, Gabriel Goh, Sandhini Agarwal, Girish Sastry, Amanda Askell, Pamela Mishkin, Jack Clark, et~al.
\newblock Learning transferable visual models from natural language supervision.
\newblock In \emph{ICML}, 2021.

\bibitem[Raffel et~al.(2020)Raffel, Shazeer, Roberts, Lee, Narang, Matena, Zhou, Li, and Liu]{raffel2020T5}
Colin Raffel, Noam Shazeer, Adam Roberts, Katherine Lee, Sharan Narang, Michael Matena, Yanqi Zhou, Wei Li, and Peter~J Liu.
\newblock Exploring the limits of transfer learning with a unified text-to-text transformer.
\newblock \emph{JMLR}, 2020.

\bibitem[Ren et~al.(2024)Ren, Jiang, Liu, Zeng, Liu, Gao, Huang, Ma, Jiang, Chen, et~al.]{ren2024grounding1.5}
Tianhe Ren, Qing Jiang, Shilong Liu, Zhaoyang Zeng, Wenlong Liu, Han Gao, Hongjie Huang, Zhengyu Ma, Xiaoke Jiang, Yihao Chen, et~al.
\newblock Grounding dino 1.5: Advance the" edge" of open-set object detection.
\newblock \emph{arXiv preprint arXiv:2405.10300}, 2024.

\bibitem[Rombach et~al.(2022)Rombach, Blattmann, Lorenz, Esser, and Ommer]{rombach2022stablediffusion}
Robin Rombach, Andreas Blattmann, Dominik Lorenz, Patrick Esser, and Bj{\"o}rn Ommer.
\newblock High-resolution image synthesis with latent diffusion models.
\newblock In \emph{CVPR}, 2022.

\bibitem[Saharia et~al.(2022)Saharia, Chan, Saxena, Li, Whang, Denton, Ghasemipour, Gontijo~Lopes, Karagol~Ayan, Salimans, et~al.]{saharia2022imagen}
Chitwan Saharia, William Chan, Saurabh Saxena, Lala Li, Jay Whang, Emily~L Denton, Kamyar Ghasemipour, Raphael Gontijo~Lopes, Burcu Karagol~Ayan, Tim Salimans, et~al.
\newblock Photorealistic text-to-image diffusion models with deep language understanding.
\newblock In \emph{NeurIPS}, 2022.

\bibitem[Salimans et~al.(2016)Salimans, Goodfellow, Zaremba, Cheung, Radford, and Chen]{salimans2016is}
Tim Salimans, Ian Goodfellow, Wojciech Zaremba, Vicki Cheung, Alec Radford, and Xi Chen.
\newblock Improved techniques for training gans.
\newblock In \emph{NeurIPS}, 2016.

\bibitem[Schuhmann et~al.(2022)Schuhmann, Beaumont, Vencu, Gordon, Wightman, Cherti, Coombes, Katta, Mullis, Wortsman, et~al.]{schuhmann2022laion5b}
Christoph Schuhmann, Romain Beaumont, Richard Vencu, Cade Gordon, Ross Wightman, Mehdi Cherti, Theo Coombes, Aarush Katta, Clayton Mullis, Mitchell Wortsman, et~al.
\newblock Laion-5b: An open large-scale dataset for training next generation image-text models.
\newblock In \emph{NeurIPS}, 2022.

\bibitem[Shirakawa and Uchida(2024)]{shirakawa2024noisecollage}
Takahiro Shirakawa and Seiichi Uchida.
\newblock Noisecollage: A layout-aware text-to-image diffusion model based on noise cropping and merging.
\newblock In \emph{CVPR}, 2024.

\bibitem[Song et~al.(2021)Song, Meng, and Ermon]{song2021ddim}
Jiaming Song, Chenlin Meng, and Stefano Ermon.
\newblock Denoising diffusion implicit models.
\newblock In \emph{ICLR}, 2021.

\bibitem[stability.ai(2024)]{sd3.5}
stability.ai.
\newblock Stable diffusion 3.5.
\newblock \url{https://stability.ai/news/introducing-stable-diffusion-3-5}, 2024.

\bibitem[Team et~al.(2024)Team, Mesnard, Hardin, Dadashi, Bhupatiraju, Pathak, Sifre, Rivi{\`e}re, Kale, Love, et~al.]{team2024gemma-paper}
Gemma Team, Thomas Mesnard, Cassidy Hardin, Robert Dadashi, Surya Bhupatiraju, Shreya Pathak, Laurent Sifre, Morgane Rivi{\`e}re, Mihir~Sanjay Kale, Juliette Love, et~al.
\newblock Gemma: Open models based on gemini research and technology.
\newblock \emph{arXiv preprint arXiv:2403.08295}, 2024.

\bibitem[Team(2024)]{omost}
Omost Team.
\newblock Omost github page.
\newblock \url{https://github.com/lllyasviel/Omost}, 2024.

\bibitem[Tian et~al.(2025)Tian, Gao, Xu, Hu, Lu, Wu, Yang, and Dehghan]{tian2025unigen}
Rui Tian, Mingfei Gao, Mingze Xu, Jiaming Hu, Jiasen Lu, Zuxuan Wu, Yinfei Yang, and Afshin Dehghan.
\newblock Unigen: Enhanced training \& test-time strategies for unified multimodal understanding and generation.
\newblock \emph{arXiv preprint arXiv:2505.14682}, 2025.

\bibitem[Wang et~al.(2024)Wang, Darrell, Rambhatla, Girdhar, and Misra]{wang2024instancediffusion}
Xudong Wang, Trevor Darrell, Sai~Saketh Rambhatla, Rohit Girdhar, and Ishan Misra.
\newblock Instancediffusion: Instance-level control for image generation.
\newblock In \emph{CVPR}, 2024.

\bibitem[Weng et~al.(2024)Weng, Huang, Qiao, Hu, Lin, Zhang, and Chen]{weng2024desigen}
Haohan Weng, Danqing Huang, Yu Qiao, Zheng Hu, Chin-Yew Lin, Tong Zhang, and CL Chen.
\newblock Desigen: A pipeline for controllable design template generation.
\newblock In \emph{CVPR}, 2024.

\bibitem[Wu et~al.(2024{\natexlab{a}})Wu, Lian, Gonzalez, Li, and Darrell]{wu2024self-correct}
Tsung-Han Wu, Long Lian, Joseph~E Gonzalez, Boyi Li, and Trevor Darrell.
\newblock Self-correcting llm-controlled diffusion models.
\newblock In \emph{CVPR}, 2024{\natexlab{a}}.

\bibitem[Wu et~al.(2024{\natexlab{b}})Wu, Zhou, Ma, Su, Ma, and Wang]{wu2024ifadapter}
Yinwei Wu, Xianpan Zhou, Bing Ma, Xuefeng Su, Kai Ma, and Xinchao Wang.
\newblock Ifadapter: Instance feature control for grounded text-to-image generation.
\newblock \emph{arXiv preprint arXiv:2409.08240}, 2024{\natexlab{b}}.

\bibitem[Xing et~al.(2023)Xing, Dai, Zhang, Zhang, Hu, Wu, and Jiang]{xing2023vidiff}
Zhen Xing, Qi Dai, Zihao Zhang, Hui Zhang, Han Hu, Zuxuan Wu, and Yu-Gang Jiang.
\newblock Vidiff: Translating videos via multi-modal instructions with diffusion models.
\newblock \emph{arXiv preprint arXiv:2311.18837}, 2023.

\bibitem[Xing et~al.(2024{\natexlab{a}})Xing, Dai, Hu, Wu, and Jiang]{xing2024simda}
Zhen Xing, Qi Dai, Han Hu, Zuxuan Wu, and Yu-Gang Jiang.
\newblock Simda: Simple diffusion adapter for efficient video generation.
\newblock In \emph{CVPR}, 2024{\natexlab{a}}.

\bibitem[Xing et~al.(2024{\natexlab{b}})Xing, Dai, Weng, Wu, and Jiang]{xing2024aid}
Zhen Xing, Qi Dai, Zejia Weng, Zuxuan Wu, and Yu-Gang Jiang.
\newblock Aid: Adapting image2video diffusion models for instruction-guided video prediction.
\newblock \emph{arXiv preprint arXiv:2406.06465}, 2024{\natexlab{b}}.

\bibitem[Xing et~al.(2024{\natexlab{c}})Xing, Feng, Chen, Dai, Hu, Xu, Wu, and Jiang]{xing2024survey}
Zhen Xing, Qijun Feng, Haoran Chen, Qi Dai, Han Hu, Hang Xu, Zuxuan Wu, and Yu-Gang Jiang.
\newblock A survey on video diffusion models.
\newblock \emph{ACM Computing Surveys}, 2024{\natexlab{c}}.

\bibitem[Xu et~al.(2023)Xu, Liu, Wu, Tong, Li, Ding, Tang, and Dong]{xu2023imagereward}
Jiazheng Xu, Xiao Liu, Yuchen Wu, Yuxuan Tong, Qinkai Li, Ming Ding, Jie Tang, and Yuxiao Dong.
\newblock Imagereward: Learning and evaluating human preferences for text-to-image generation.
\newblock In \emph{NeurIPS}, 2023.

\bibitem[Xue et~al.(2023)Xue, Huang, Sun, Song, and Zhang]{xue2023freestyle}
Han Xue, Zhiwu Huang, Qianru Sun, Li Song, and Wenjun Zhang.
\newblock Freestyle layout-to-image synthesis.
\newblock In \emph{CVPR}, 2023.

\bibitem[Yang et~al.(2023)Yang, Luo, Chen, Wang, Liang, and Lin]{yang2023law-diffusion}
Binbin Yang, Yi Luo, Ziliang Chen, Guangrun Wang, Xiaodan Liang, and Liang Lin.
\newblock Law-diffusion: Complex scene generation by diffusion with layouts.
\newblock In \emph{ICCV}, 2023.

\bibitem[Yang et~al.(2024)Yang, Yu, Meng, Xu, Ermon, and Bin]{yang2024rpg-master}
Ling Yang, Zhaochen Yu, Chenlin Meng, Minkai Xu, Stefano Ermon, and CUI Bin.
\newblock Mastering text-to-image diffusion: Recaptioning, planning, and generating with multimodal llms.
\newblock In \emph{ICML}, 2024.

\bibitem[Ye et~al.(2023)Ye, Zhang, Liu, Han, and Yang]{ye2023ipadapter}
Hu Ye, Jun Zhang, Sibo Liu, Xiao Han, and Wei Yang.
\newblock Ip-adapter: Text compatible image prompt adapter for text-to-image diffusion models.
\newblock \emph{arXiv preprint arXiv:2308.06721}, 2023.

\bibitem[Zheng et~al.(2023)Zheng, Zhou, Li, Qi, Shan, and Li]{zheng2023layoutdiffusion}
Guangcong Zheng, Xianpan Zhou, Xuewei Li, Zhongang Qi, Ying Shan, and Xi Li.
\newblock Layoutdiffusion: Controllable diffusion model for layout-to-image generation.
\newblock In \emph{CVPR}, 2023.

\bibitem[Zhou et~al.(2024)Zhou, Li, Ma, Zhang, and Yang]{zhou2024migc}
Dewei Zhou, You Li, Fan Ma, Xiaoting Zhang, and Yi Yang.
\newblock Migc: Multi-instance generation controller for text-to-image synthesis.
\newblock In \emph{CVPR}, 2024.

\end{thebibliography}
}
\clearpage
\setcounter{page}{1}
\maketitlesupplementary

\renewcommand{\thesection}{\Alph{section}}
\setcounter{section}{0}

\section{More Details on SiamLayout-FLUX}
\begin{figure}[h]
  \centering
  \includegraphics[width=1.0\linewidth]{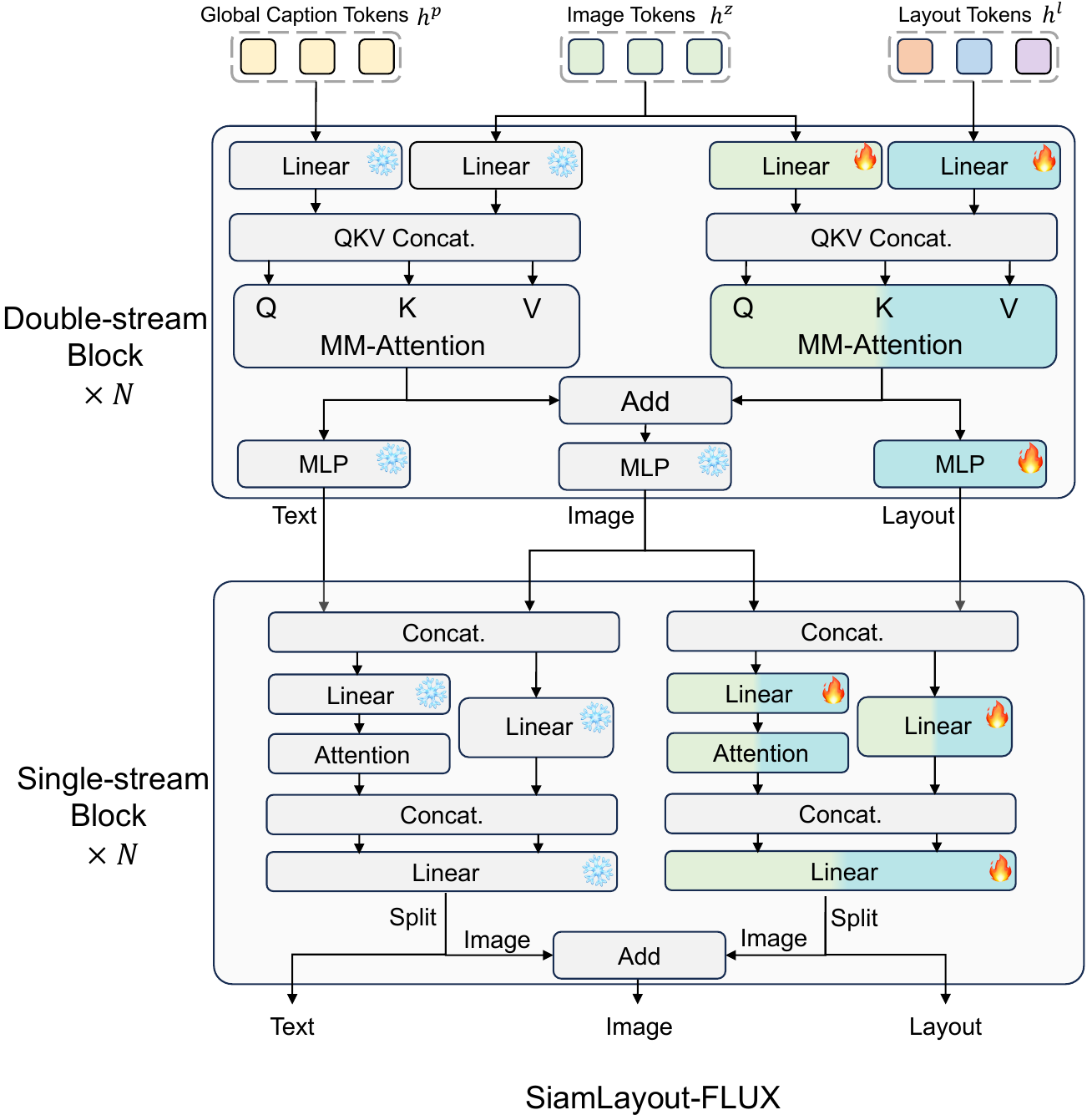}
  \caption{An overview of SiamLayout-FLUX.}
  \label{fig:SiamLayout_FLUX}
\end{figure}

FLUX is another outstanding text-to-image generative model based on the Multimodal Diffusion Transformer (MM-DiT) architecture, demonstrating remarkable performance alongside SD3. The core components of FLUX include the double-stream block and the single-stream block. To further verify the generality of our proposed SiamLayout approach, we integrate SiamLayout into FLUX and propose SiamLayout-FLUX, aiming to empower FLUX for layout-to-image generation.

Specifically, as illustrated in \cref{fig:SiamLayout_FLUX}, for the Double Stream Block in FLUX, we adopt the same layout integration strategy as presented in \cref{fig:architecture}. For the Single Stream Block, we maintain the core concept from SiamLayout, which emphasizes preserving the advantage of MM-Attention in facilitating effective multimodal interactions while reducing interference and competition among modalities. Therefore, we similarly introduce a Siamese branch dedicated to processing the interactions between images and layouts. Such a design enables the layout information to independently and concurrently guide the image content generation, analogous to the role played by the global text caption.

Both quantitative and qualitative experimental results demonstrate that our proposed SiamLayout-FLUX outperforms previous methods by a clear margin, validating the effectiveness of SiamLayout and its general applicability across different MM-DiTs.

\section{LoRA Version of SiamLayout}
\begin{figure}[h]
  \centering
  \includegraphics[width=1.0\linewidth]{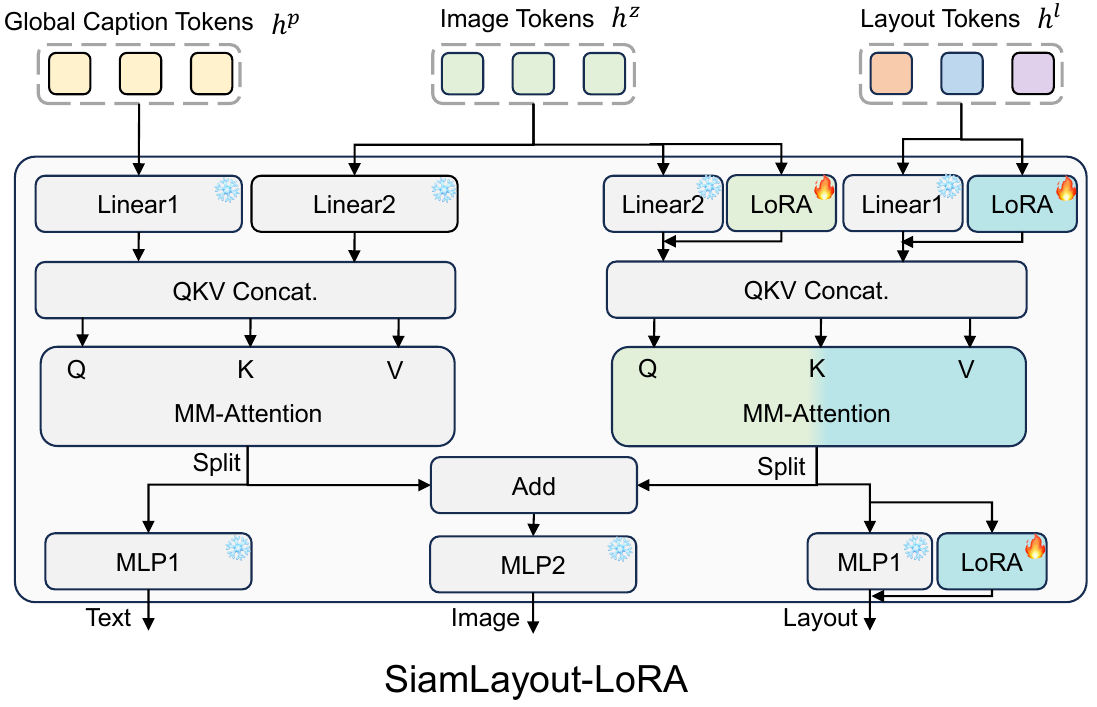}
  \caption{An overview of SiamLayout-LoRA.}
  \label{fig:SiamLayout_LoRA}
\end{figure}
We further propose a more lightweight variant for layout control, named SiamLayout-LoRA, which achieves comparable layout control accuracy with fewer additional parameters. As illustrated in \cref{fig:SiamLayout_LoRA}, SiamLayout-LoRA integrates Low-Rank Adaptation (LoRA) modules into the Linear1, Linear2, and MLP1 layers, efficiently handling image-layout interactions based on the frozen pre-trained weights. This LoRA-based variant provides alternative trade-offs between extra parameter overhead and layout control accuracy. In our experiments, we set the LoRA rank to 256.

\section{Discussion on Extra Parameters and Computation Introduced by Layout Control}

\begin{table}[h]
\centering
\tabcolsep=0.05cm
\ra{1.1}
\scalebox{0.63}{
\begin{tabular}{@{}lcccccc@{}}
\toprule
                               & Extra Params~$\downarrow$ & Extra MACs~$\downarrow$ & Spatial~$\uparrow$ & Color~$\uparrow$ & Texture~$\uparrow$ & Shape~$\uparrow$ \\ \midrule
GLIGEN                         & 24.20\%      & 30.80\%    & 77.53   & 49.41 & 55.29   & 52.72 \\
InstanceDiff                   & 42.67\%      & 155.83\%   & 87.99   & 69.16 & 72.78   & 71.08 \\
HiCo                           & 42.03\%      & 263.70\%   & 87.04   & 69.19 & 72.36   & 71.1  \\ \midrule
Layout Adapter                 & 16.90\%      & \underline{21.70\%}    & 88.43   & 71.67 & 73.56   & 72.61 \\
$\mathrm{M}^3$-Attention       & 48.50\%      & \textbf{15.70\%}    & 79.15   & 60.19 & 62.96   & 61.29 \\ \midrule
SiamLayout-SD3                     & 62.40\%      & 38.70\%    & \textbf{92.67}   & \textbf{74.45} & \textbf{77.21}   & \textbf{75.93} \\
SiamLayout-SD3-LoRA                & \textbf{15.90\%}      & 39.90\%    & \underline{89.71}   & \underline{71.89} & \underline{74.14}   & \underline{72.63} \\ \bottomrule
\end{tabular}}
\caption{Comparison of extra parameters and computation. \textbf{Bold}, \underline{underline} represent the best and second best methods, respectively.}
\label{tab: extra parameters and computation}
\end{table}

As shown in \cref{tab: extra parameters and computation}, we compare different approaches with respect to extra parameters and computational overhead (i.e., MACs required for one denoising step given a layout containing 10 entities). To ensure a fair comparison, we report the relative increase in parameters and computation introduced by integrating layout control compared to the base model.

Our proposed approach outperforms others at comparable additional parameter counts while requiring significantly lower extra computational overhead compared to InstanceDiff and HiCo. This is due to InstanceDiff and HiCo computing image-layout attention separately for each entity before integrating them, causing their computational overhead to increase linearly as the number of entities grows.
In addition, our proposed LoRA variant achieves comparable layout control accuracy with substantially fewer extra parameters.

\section{More Details on Datasets and Benchmarks}

\subsection{LayoutSAM Dataset and Benchmark}

\paragraph{Layout annotation pipeline.}
We design a mechanism to automatically annotate the layout for any given image.

\begin{figure}[h]
  \centering
  \includegraphics[width=1.0\linewidth]{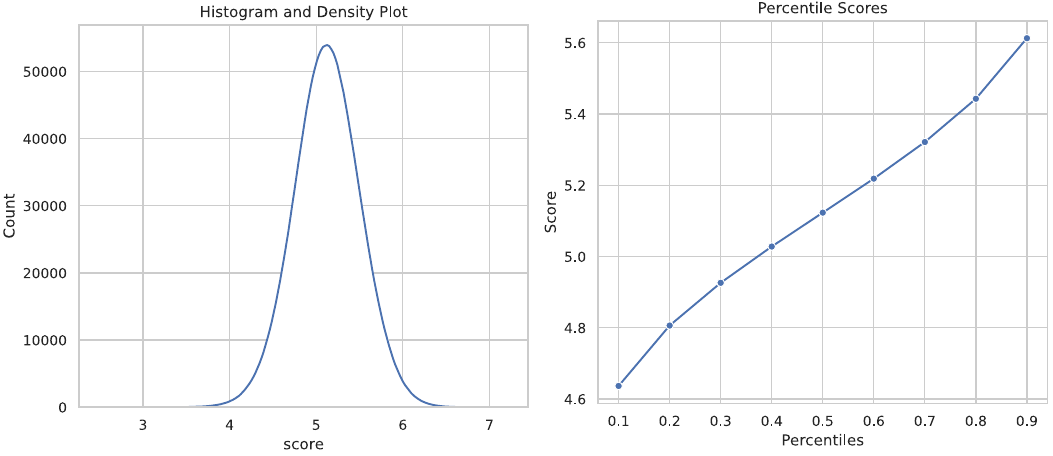}
  \caption{Distribution of aesthetic scores of SAM.}
  \label{fig:supp_data_statistic}
\end{figure}

\Rmnum{1}) Image Filtering: We employ the LAION-Aesthetics predictor~\cite{LAION-Aesthetics} to assign aesthetic scores to images and filter out those with low scores. For SAM~\cite{kirillov2023sam}, we analyze the aesthetic scores shown in \cref{fig:supp_data_statistic} and curate a high visual quality subset consisting of images in the top 50\% of aesthetic scores.

\Rmnum{2}) Global Caption Annotation: We generate the detailed descriptions for the query image using the Qwen-VL-Chat-Dense-Captioner~\cite{Qwen-VL-Captioner}, which is a vision-language model fine-tuned on generative and human-annotated data using LoRA. This model supports accurate and detailed image descriptions. The average length of the captions is 95.41 tokens, measured by the CLIP tokenizer.

\Rmnum{3}) Entity Extraction: Existing state-of-the-art open-set grounding models~\cite{liu2023grounding1,ren2024grounding1.5} perform better at detecting entities using a list of short phrases than directly using dense captions. Thus, we utilize the large language model Llama3.1-8b-it~\cite{llama3.1} to extract the main entities from dense captions via in-context learning. The brief descriptions include simple attribute descriptions with an average length of 2.08 tokens.

\Rmnum{4}) Entity Spatial Annotation: We use Grounding DINO~\cite{liu2023grounding1} to annotate bounding boxes of entities. To clean noisy data, we design the following filtering rules. We first filter out bounding boxes that occupy less than 2\% of the total image area, then only retain images with 3 to 10 bounding boxes. The average number of entities per image is 3.96.

\Rmnum{5}) Region Caption Recaptioning: We use the vision language model MiniCPM-V-2.6~\cite{minicpm-v-2.6} to generate fine-grained descriptions with complex attributes for each entity based on its visual content and brief description. The generated detailed descriptions generally cover attributes such as color, shape, texture, and some text details, with an average length of 15.07 tokens. 

\begin{figure*}[]
  \centering
  \includegraphics[width=1.0\linewidth]{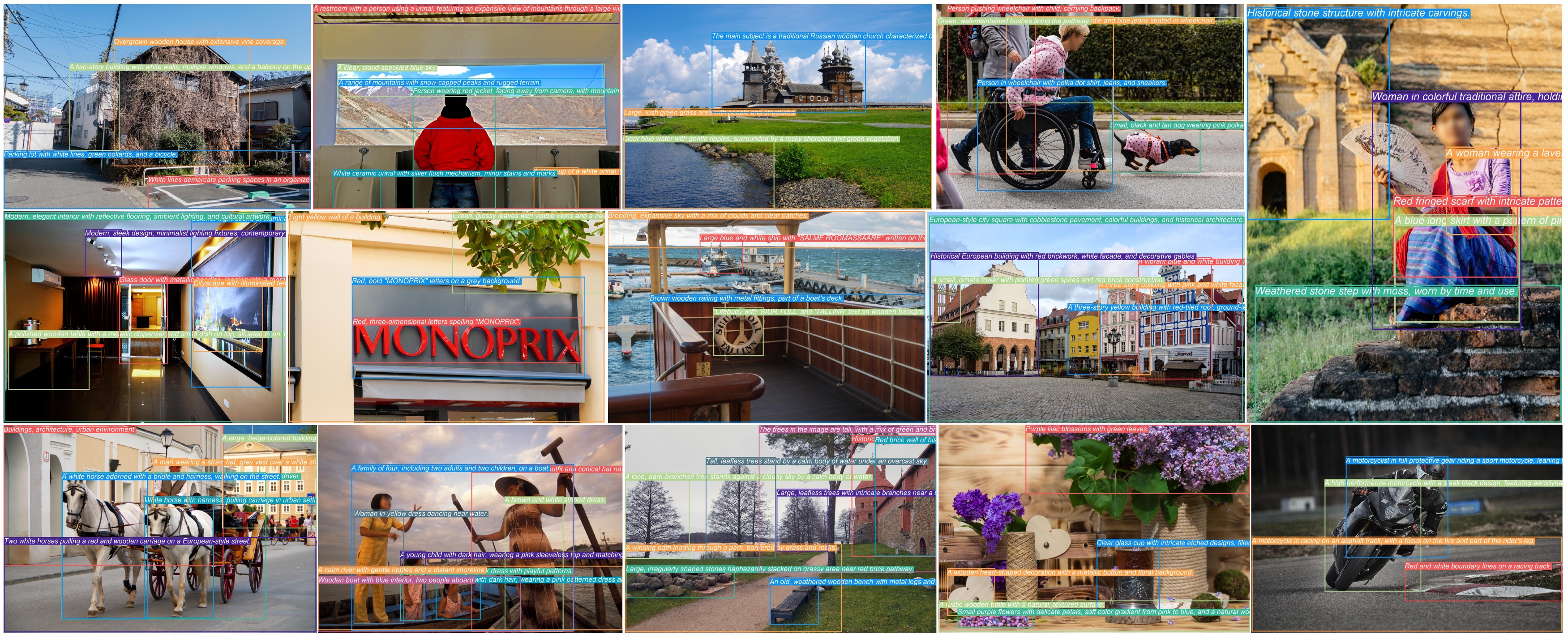}
  \caption{Examples from the LayoutSAM dataset.}
  \label{fig:supp_data_samples}
\end{figure*}

Finally, we contribute the large-scale layout dataset LayoutSAM, which includes 2.7 million image-text pairs and 10.7 million entities. Each entity is annotated with a bounding box and a detailed description. \cref{fig:supp_data_samples} shows some examples from LayoutSAM.

\paragraph{LayoutSAM-Eval Benchmark.}
The LayoutSAM-Eval benchmark, constructed from LayoutSAM, serves as a comprehensive tool for evaluating layout-to-image generation quality. It consists of 5,000 layout data points.
We evaluate layout-to-image generation quality using LayoutSAM-Eval from two aspects: Spatial and Attribute, both of which are evaluated via the vision language model MiniCPM-V-2.6~\cite{minicpm-v-2.6} in a visual question-answering manner.
\begin{itemize}
\item[--] \textit{Spatial Accuracy.} To measure spatial adherence, for each bounding box, we ask the VLM whether the given entity exists within the bounding box, with the answer being either ``Yes'' or ``No.'' Finally, we divide the number of entities with a ``Yes'' answer by the total number of entities to obtain the spatial score.
\item[--] \textit{Attribute Accuracy.} To measure attribute adherence, we ask the VLM whether the entity within the bounding box matches the attributes in the detailed description. For attributes like color, shape, and texture, each attribute is evaluated independently through visual question answering, and the score is obtained in the same manner as the spatial score.
\end{itemize}

\subsection{Layout Planning Dataset and Benchmark}
\paragraph{Layout Planning Dataset.}
To train LayoutDesigner, we construct a layout planning dataset derived from LayoutSAM. It consists of a total of 180,000 data points, covering the following three tasks, each with 60,000 data points:
\begin{itemize}
\item[--] \textit{Caption-to-layout generation.} We randomly select data from LayoutSAM to construct pairs of global captions and ground truth layouts of entities. Each entity includes a bounding box and a description. This portion of the data is used to train the generation of layouts based on global captions.

\item[--] \textit{Center point-to-layout generation.} For each entity, we calculate the center point from its bounding box to construct pairs of center points and ground truth layouts of entities. This portion of the data is used to train the generation of layouts based on the center points of entities.

\item[--] \textit{Suboptimal layout-to-layout optimization.} For each entity, we create suboptimal layouts by performing operations such as deletion, duplication, movement, and resizing with a certain probability. This portion of the data consists of pairs of suboptimal layouts and GT layouts and is used to train the optimization of layouts from suboptimal to better.
\end{itemize}

LayoutDesigner is a unified layout planning model that supports all three tasks simultaneously through joint training of a large language model on these three types of data.

\paragraph{Layout Planning Benchmark.}
We construct 1,000 data points for each layout planning task using the same method from LayoutSAM-Eval and conduct experiments to evaluate layout planning capabilities under a 3-shot in-context learning setting.
First, we evaluate the formatting accuracy of the generated layouts, including the coordinates of the top-left corner being smaller than those of the bottom-right corner and the bounding box not exceeding the image boundaries.
To assess the harmony and aesthetics of the layouts, we did not use metrics like AP to measure the adherence of the generated bounding boxes to the ground truth. This is because layout generation is an open-ended problem, and there can be multiple optimal solutions for the input. Even if a solution does not resemble the GT, it can still be an excellent layout.
Therefore, we generate images based on the designed layouts and reflect the quality of the layouts through the quality of the images. Additionally, we evaluate the layout planning capabilities through qualitative results.

\section{More Analysis on Modal Competition}
\begin{figure}[h]
  \centering
  \includegraphics[width=1.0\linewidth]{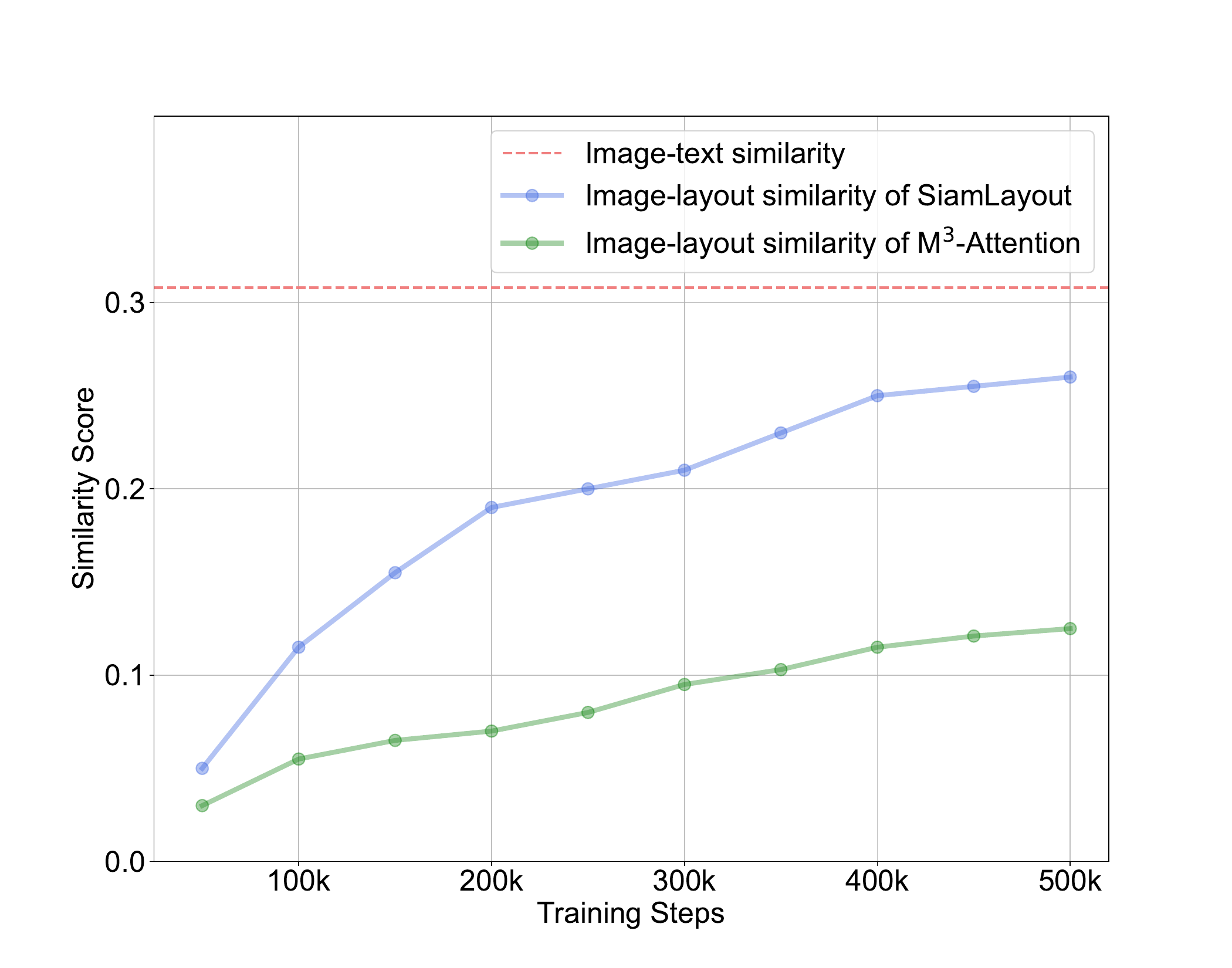}
  \caption{The trend of image-layout similarity in the attention maps of different network variants.}
  \label{fig:supp_modal_competition_visual}
\end{figure}
In \cref{fig:Modality_competition}, we analyze the issue of modality competition in $\mathrm{M}^3$-Attention: the similarity between the layout and the image is much lower than that between the global caption and the image. Our proposed SiamLayout alleviates this issue by decoupling it into two siamese MM-Attention branches: image-text and image-layout. 
We investigate the image-text and image-layout similarity scores in the attention map to further illustrate the modality competition, as shown in ~\cref{fig:supp_modal_competition_visual}.
The image-layout similarity score is determined by applying softmax to the attention map of the $\mathrm{M}^3$-Attention or image-layout MM-Attention, identifying the cross-region of the image and layout, and then taking the average of the top 1\% scores in this cross-region. The image-text similarity score is calculated in the same way.
Experimental results show that, as the training steps increase, the image-layout similarity score in $\mathrm{M}^3$-Attention remains consistently much lower than the image-text similarity score, resulting in the layout having a weaker influence on the image compared to the global caption.
In contrast, due to the independent guiding role of the layout in the image-layout branch of SiamLayout, the image-layout similarity gradually increases to a value close to the image-text similarity as the network training progresses, thereby allowing the layout to play a more significant guiding role in image generation.

\section{More Qualitative Results}
\begin{figure*}[]
  \centering
  \includegraphics[width=1.0\linewidth]{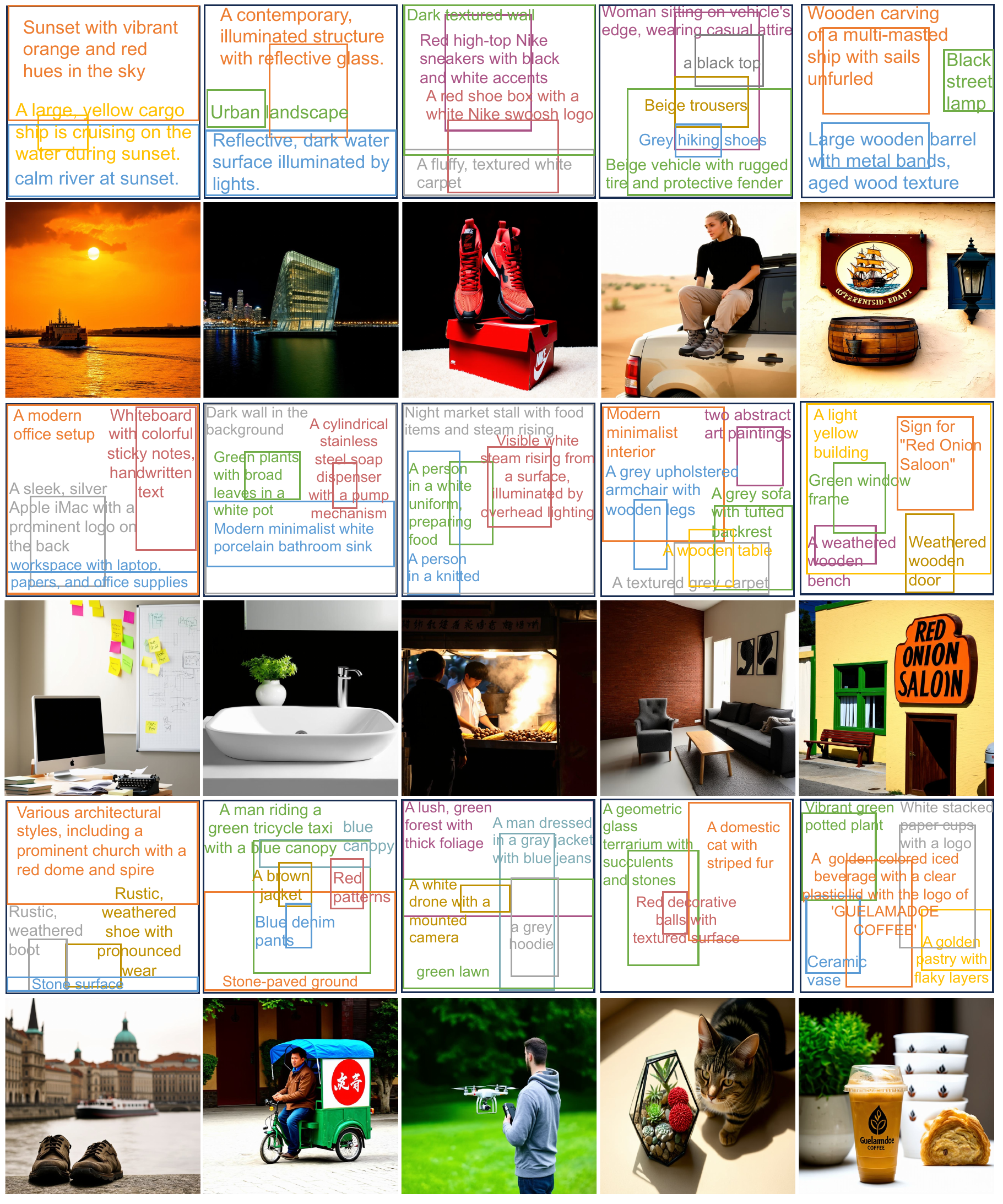}
  \caption{More qualitative results on layout-to-image generation.}
  \label{fig:supp_more_visualization}
\end{figure*}

We present more qualitative results of SiamLayout-SD3 in \cref{fig:supp_more_visualization}. Experimental results show that our proposed method empowers MM-DiT for layout-to-image generation, achieving visually appealing and precisely controllable generation, as demonstrated by the high adherence to complex attributes such as color, texture, and shape.

\end{document}